\documentclass{article}
\usepackage{iclr2027_conference,times}

\usepackage{amsmath,amsfonts,bm}

\def\eqref#1{equation~\ref{#1}}
\def\1{\bm{1}}

\DeclareMathAlphabet{\mathsfit}{\encodingdefault}{\sfdefault}{m}{sl}
\SetMathAlphabet{\mathsfit}{bold}{\encodingdefault}{\sfdefault}{bx}{n}

\usepackage{tabularx}
\usepackage{array}
\usepackage{amsmath}
\usepackage{amssymb}
\usepackage{bbm}
\usepackage{booktabs}
\usepackage{graphicx}
\usepackage{hyperref}
\usepackage{multirow}
\usepackage{url}
\usepackage{soul}
\usepackage[table]{xcolor}
\usepackage{graphicx}
\usepackage{ragged2e}
\usepackage{placeins}
\usepackage{float}
\usepackage[most]{tcolorbox}
\usepackage{booktabs}
\usepackage{algorithm}
\usepackage{algpseudocode}

\title{Geospatial Vector Geometry as a Native Spatial Modality for Vision-Language Models}

\author{
Huang Fei$^{1}$, 
Boyi Chen$^{1}$, 
Hao Xue$^{1}$, 
Yang Yue$^{1*}$, 
Yehong Xu$^{2}$, 
Xiaofang Zhou$^{2}$ \\[0.6em]
\normalfont
$^{1}$The Hong Kong University of Science and Technology (Guangzhou), China \\
$^{2}$The Hong Kong University of Science and Technology, Hong Kong SAR, China \\
$^{*}$Corresponding authors
}

\iclrfinalcopy

\begin{document}
\raggedbottom

\maketitle
\lhead{Preprint}

\begin{abstract}
Geospatial reasoning links VLMs to real-world action and decision making. Reliable reasoning requires accurate spatial information and effective spatial representations. However, current VLMs obtain this information mainly from language and vision. At larger geographic scales, limited visual observations make it difficult to establish accurate spatial references and recover precise spatial relations. This leads to a bottleneck for VLM spatial perception and reasoning. This paper proposes that \textit{geospatial vector geometry can become a native spatial modality within a VLM} and introduces GeoVec-VLM. First, we develop a structured geometry representation that preserves the spatial structure and arrangement of vector objects, building an explicit spatial basis. Second, we introduce a geometry interaction strategy that brings this basis into multimodal computation, where it interacts with visual content and language references according to the queried spatial relations. Together, these designs allow perception and reasoning to operate on shared geometric references. Across five spatial reasoning tasks, GeoVec-VLM improves mean accuracy by about 11\% relative to the strongest baseline, maintains its gains under geographic transfer, and shows greater geometric consistency. Our results show that geospatial vector geometry offers a promising internal spatial basis for VLM spatial reasoning. \url{https://github.com/TXFR/GeoVector-VLM}
\end{abstract}

\section{Introduction}
Spatial intelligence is increasingly expected to understand and reason in geographic space, from navigation and autonomous driving to map-based spatial decision making \citep{anderson2018vision,guo2025surds,dihan2025mapeval}. Recent Vision-Language Models (VLMs) have made rapid progress in spatial reasoning within visual and 3D environments  \citep{yang2025thinking,wang2026mindcube}, where camera-centered observations provide rich cues for recovering local geometry and relations among nearby objects, e.g., a chair is meters from a door. However, reasoning in geographic space is more difficult as it often spans vast extents, from hundreds of meters to tens of kilometers, while still requiring precise geometric relations \citep{stevens1978distortions,dihan2025mapeval}. Under a fixed visual input and encoding budget, these spatial structures remain difficult to recover accurately from visual observations alone \citep{liu2018coordconv,rahmanzadehgervi2024vision,chen2024spatialvlm}.  External tools can perform precise spatial computation \citep{zhang2024geogpt,zhang2025geospatial}, but they do not directly change spatial representation inside the model. We ask: \emph{how should spatial structure enter a VLM to support reliable geographic-scale reasoning?} We consider it from two perspectives:

(i) Recent advances have begun to introduce visual primitives \citep{chen2023shikra,fan2025grit,lu2026thinking} such as points and bounding boxes that bring spatial references directly into the model's computation, improving spatial grounding and perception \citep{xu2024pixelaligned,cheng2026grounded}. For geospatial reasoning, however, these location- and region-level primitives are insufficient, especially for representing object geometry and spatial layout. (ii) Vector data naturally provide such spatial structure. Widely used in geographic information systems and spatial databases, they encode geometry and metric position in a shared coordinate system, supporting precise spatial measurements and relational computations \citep{guting1994introduction,jeung2010path,zheng2012spatial,saeedan2026applicability}. Yet current interfaces often either serialize coordinates into language-like sequences \citep{ji2025foundation} or compress spatial objects into fixed-size embeddings \citep{mai2023polygon,siampou2025poly2vec,chu2026geo2vec}, weakening the spatial organization available for multimodal reasoning \citep{knoblauch2026beyondpixels}.

Driven by this analysis, we introduce \textbf{GeoVec-VLM}, which incorporates \emph{\textbf{Geo}spatial \textbf{Vec}tor data as a native spatial modality for VLMs}. GeoVec-VLM addresses two challenges: preserving the spatial structure of vector data inside the model, and enabling this structure to interact with vision and language to provide a spatial bias for reasoning. First, GeoVec-VLM encodes vector objects as point and path tokens that preserve both global arrangement and local geometry, providing a geometric basis for spatial relations such as direction and topology. Second, inside the VLM, geometric biases and learnable spatial states bring this structure into multimodal computation, associating vector geometry with visual content and language references while forming query-specific spatial relations for reasoning and language generation. Structured spatial objectives supervise these states to be verifiable during training. We build a large-scale dataset of geospatial reasoning examples, and extensive experiments show consistent gains over images and serialized-geometry interfaces and strong generalization under zero-shot geographic transfer. Beyond accuracy, GeoVec-VLM exhibits stronger geometric consistency across related spatial judgments. Our contributions are as follows:

\begin{itemize}
    \item We introduce geospatial vector geometry as a native spatial modality for VLMs, allowing its spatial structure to form an internal basis for organizing multimodal spatial reasoning.

    \item We propose GeoVec-VLM, which uses a structure-preserving vector representation and joint vector–vision–language association to connect vector structure with visual content, allowing geometric information to guide multimodal spatial reasoning and language generation.

    \item Our findings show that language-level accuracy can mask geometrically inconsistent spatial reasoning in VLMs. Native vector integration improves accuracy and better aligns model reasoning with the underlying spatial constraints.
\end{itemize}

\begin{figure}[t]
\centering
\includegraphics[width=\linewidth]{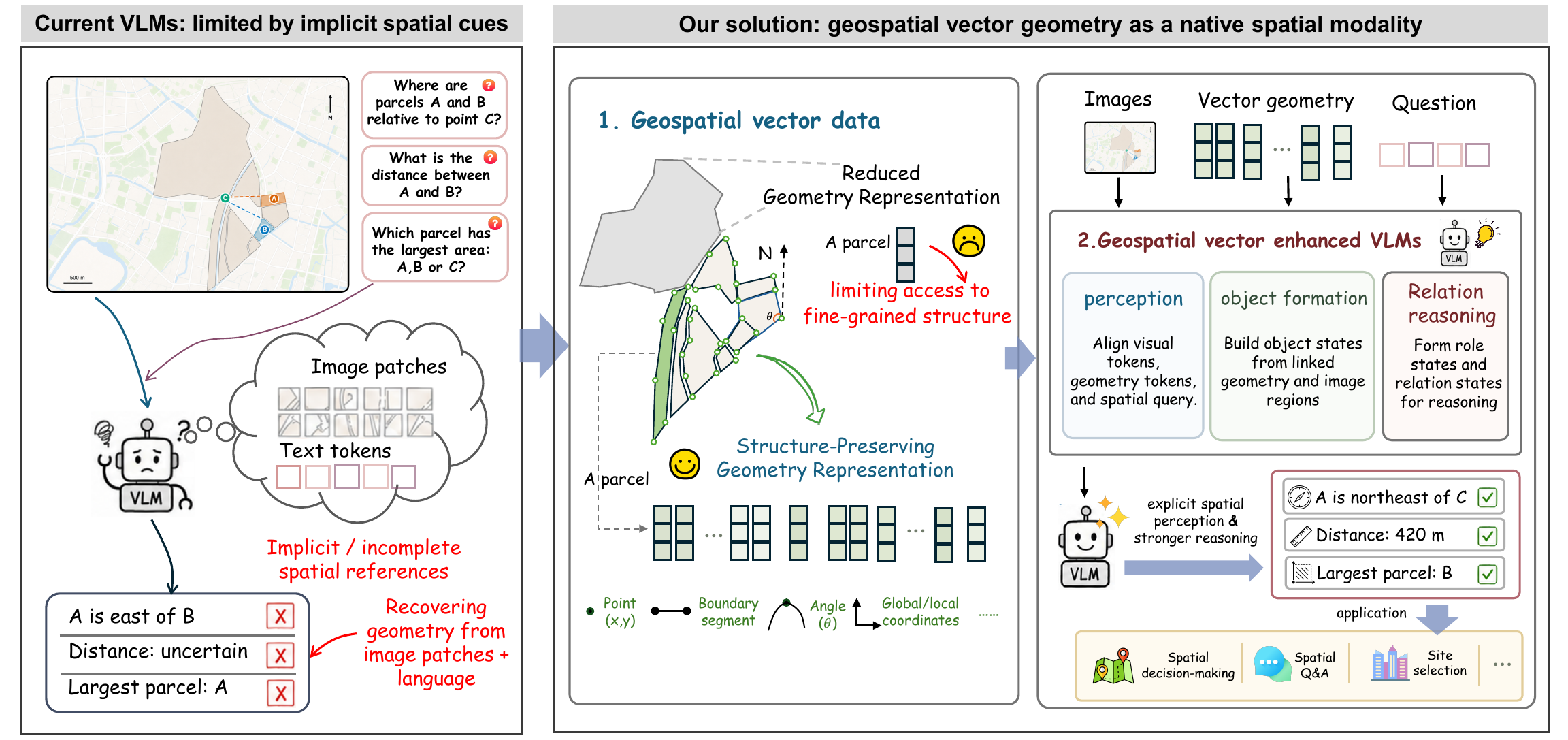}
\setlength{\abovecaptionskip}{0pt}
\caption{Motivation and overview of GeoVec-VLM. Current VLMs infer spatial relations from raster images and language, where metric geometry and spatial relations are only implicitly represented. GeoVec-VLM introduces geospatial vector data as a native spatial modality, preserves its geometric structure during representation, and integrates it with visual and language information for spatial perception and reasoning.}
\label{fig:tasks}
\end{figure}

\section{Related Work}

\subsection{Spatial Reasoning and Spatial Primitives}

Spatial reasoning has been studied in both large language models (LLMs) and vision-language models. LLM-based work has examined geographic knowledge and reasoning across global and urban settings \citep{manvi2024geollm, feng2025citygpt}. More recent studies probe topological reasoning either from natural-language
queries or by providing serialized vector geometries directly to LLMs \citep{xie2025topological, ji2025foundation}. In parallel, VLM-based approaches improve metric and grounded spatial reasoning through spatial supervision, region-level representations, and depth cues \citep{chen2024spatialvlm, cheng2024spatialrgpt}, while recent benchmarks continue to reveal limitations in fine-grained and compositional spatial reasoning \citep{jia2026omnispatial}. Recent studies further suggest that spatial reasoning cannot be addressed solely by stronger linguistic reasoning. For example, \citet{yang2025thinking} find that standard linguistic reasoning strategies provide limited gains on visual-spatial tasks, whereas explicitly constructed cognitive maps improve spatial distance reasoning. A related line of work makes spatial information explicit during multimodal reasoning. Visual CoT \citep{shao2024visualcot} grounds intermediate reasoning steps in image regions, GRIT interleaves language reasoning with bounding-box coordinates \citep{fan2025grit}, and Thinking with Visual Primitives \citep{lu2026thinking} uses points and boxes as reasoning primitives, including sequences of points for spatial paths. These approaches demonstrate the value of keeping spatial references accessible during reasoning. Their spatial interfaces, however, primarily operate on points, regions, or bounding boxes. Polygon geometries additionally carry metric and relational structure, including ordered coordinates, path progression, local direction, and object extent---from which spatial relations can be derived. Representing this structure while making it available to multimodal reasoning introduces a different representation problem.

\subsection{Vector Geometry for Multimodal Spatial Reasoning}

Vector geometry provides spatial structure at multiple levels, from coordinates
and local shape to paths, boundaries, and object extent. Existing spatial
encoders capture this structure in increasingly expressive ways. Polygon
encoders model vertex configurations and geometric properties
\citep{mai2023polygon}, while Poly2Vec and Geo2Vec extend geometry-aware
encoding to points, polylines, and polygons
\citep{siampou2025poly2vec, chu2026geo2vec}. These methods preserve rich
geometry during encoding, but typically expose each spatial object to
downstream models through a fixed-size representation. Recent work has further connected geometric information with language and
vision. LLMs can reason over serialized vector geometries
\citep{ji2025foundation}, while SLM \citep{chu2026slm} projects Geo2Vec representations into the
language-model embedding space. Rose \citep{bai2025rose} and GeoLink \citep{bai2025geolink} combine imagery with vector map data, allowing image patches to
interact with vector objects. Beyond geospatial vector data, recent spatial VLMs
also integrate geometric features with visual and language representations
through multimodal fusion \citep{zhang2026spatialstack,zhao2026spacemind}.
These studies show that geometric information can improve multimodal spatial
reasoning, while also highlighting the interface between geometry, vision, and
language as an important modeling choice. For complex vector objects, however,
geometry is generally summarized at the object level before such multimodal
interaction. Recent discussions of raster--vector learning similarly highlight
the need to retain fine-grained vector structure across modalities
\citep{knoblauch2026beyondpixels}. GeoVec-VLM keeps the metric spatial reference and structure of vector objects accessible during vector-vision-language interaction. This provides a spatial basis through which visual content, language references, and
vector geometry can be associated within the same spatial reasoning process.

\section{Method}

\subsection{Problem Formulation and Overview}
\label{sec:overview}

GeoVec-VLM treats vector geometry as a native spatial modality alongside vision and language. Formally, given a raster image $I$, vector geometry $X$ in the same spatial frame, and a language instruction $T$ specifying the spatial reasoning task, the model generates an answer $y$. We write it as:
\begin{equation}
G=\mathcal{E}_{\mathrm{vec}}(X), \qquad
y=\mathcal{M}_{\mathrm{VLM}}\!(
\underbrace{I,T,G,Q}_{\text{\scriptsize model inputs}}
\,;\,
\underbrace{O,A,Z}_{\text{\scriptsize spatial states}}
).
\label{eq:framework}
\end{equation}
where $G$ denotes the structure-preserving geometry tokens, and $Q$ is included only for tasks with explicit numerical spatial inputs. Three types of learnable spatial states organize how geometry interacts with vision and language during multimodal reasoning.

\begin{figure}[t]
\centering
\includegraphics[width=\linewidth]{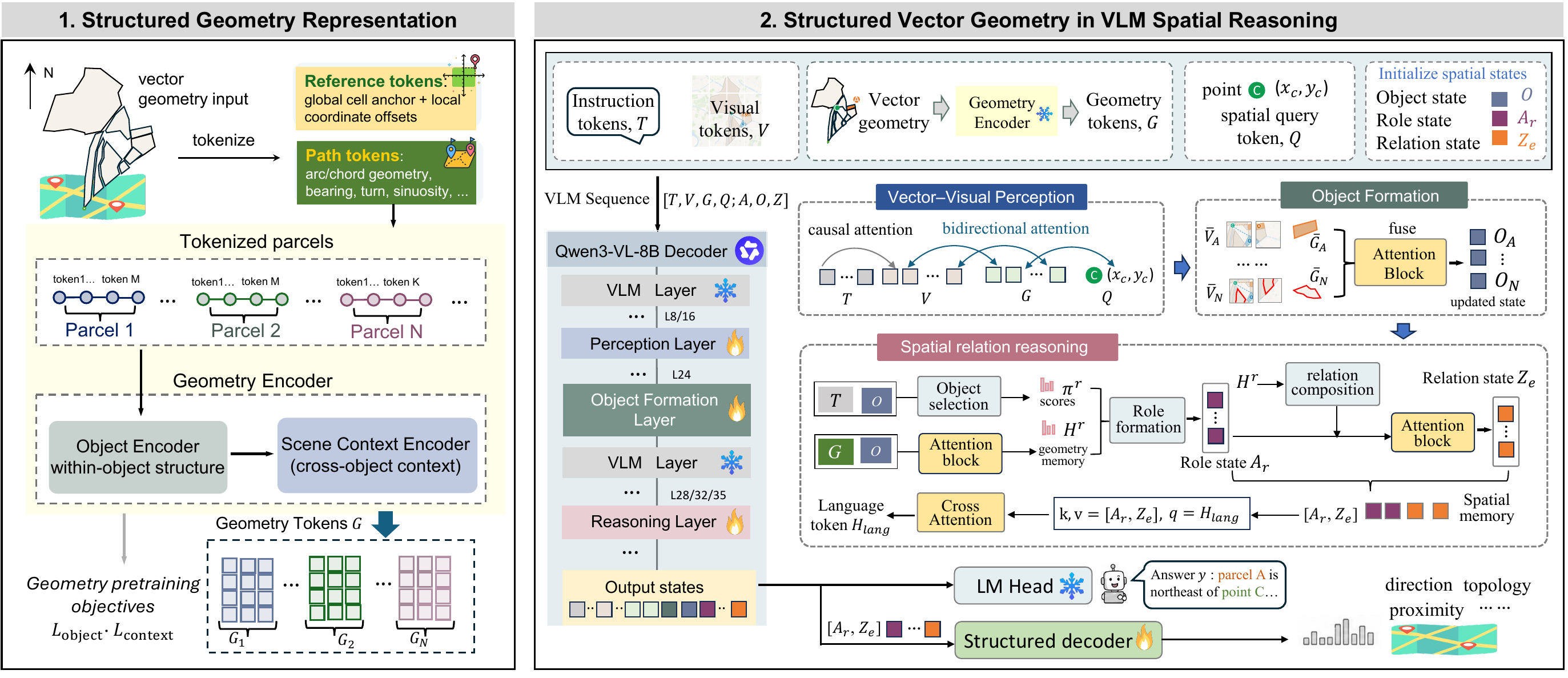}
\setlength{\abovecaptionskip}{-1pt}
\caption{Architecture of GeoVec-VLM. The geometry encoder preserves within-object structure and scene-level metric structure in geometry tokens. Inside the VLM, geometry tokens interact with visual and spatial-query tokens, are organized into object states, and form query-conditioned role and relation states used by the language generation path. Structured decoders provide spatial supervision during training.}
\label{fig:framework}
\end{figure}

As shown in Fig.~\ref{fig:framework} and Eq.~\ref{eq:framework}, GeoVec-VLM follows two main steps. First, a structured geometry encoder transforms vector geometry into geometry tokens while preserving both object geometry and scene-level spatial organization (Section~\ref{sec:geometry-encoding}). Second, GeoVec-VLM brings these geometry tokens directly into multimodal computation using spatial states inside the VLM. Section~\ref{sec:integration} associates vector geometry with visual contexts via object states $O$. Section~\ref{sec:reasoning} uses role states $A$ and relation states $Z$ to organize query-relevant geometry for spatial reasoning and language generation.

\subsection{Structure-Preserving Vector Geometry Encoding}
\label{sec:geometry-encoding}
Spatial structure in vector data is determined by object geometry and georeferenced position. Geometry defines shape, boundary orientation, and local characteristics, while position dictates direction, distance, and overall spatial arrangement. Our vector encoder preserves both forms of structure when converting vector data into geometry tokens.

Consider a scene with $N$ polygonal geometries $X=\{X_i\}_{i=1}^{N}$, where $X_i$ denotes the $i$-th object. We represent each polygon boundary as a variable-length ordered sequence of reference points and path segments, $X_i = (x_{i,j})_{j=1}^{M_i}$, where $M_i$ is the number of geometric elements for geometry $i$. Each geometric element $x_{i,j}$ is associated with its source object $\mu_i$, normalized arc position $\sigma_{i,j}$ along the boundary, element type $\tau_{i,j}$, and metric coordinate $c_{i,j}\in\mathbb{R}^2$. Continuous features $f_{i,j}$ describe its local geometry: $x_{i,j}=\big(f_{i,j},\mu_i,\sigma_{i,j},\tau_{i,j},c_{i,j}\big)$. This representation keeps object membership, boundary order, and metric position explicit. 

We encode the resulting geometry at two spatial scales. Within each polygon, let $\sigma_i=(\sigma_{i,j})_{j=1}^{M_i}$  and $\tau_i=(\tau_{i,j})_{j=1}^{M_i}$ denote the corresponding arc-position and element-type sequences. An object encoder models the ordered boundary using the relative arc position of its geometric elements:
\begin{equation}
\widetilde U_i
=
E_{\mathrm{obj}}
\left(
X_i;
PE_{\mathrm{arc}}(\sigma_i),
\tau_i
\right).
\end{equation}
We then jointly encode all polygons in the shared scene coordinate system. Using their metric coordinates $C=\{c_{i,j}\}$ and source-parcel identities, a scene encoder models the spatial configuration across parcels:
\begin{equation}
U
=
E_{\mathrm{scene}}
\left(
\{\widetilde U_i\}_{i=1}^{N};
\operatorname{RoPE}_{2D}(C)
\right),
\qquad
U=\{U_i\}_{i=1}^{N},\quad
U_i=(u_{i,j})_{j=1}^{M_i}.
\end{equation}
The first stage models geometric structure within each object, while the second models metric relations across objects. Its output remains element-wise, with each input element $x_{i,j}$ represented by a scene-contextualized state $u_{i,j}$ rather than being pooled at the parcel level. Each $u_{i,j}$ is projected into the VLM hidden space as a geometry token $g_{i,j}$. The resulting geometry tokens are $G=\{G_i\}_{i=1}^{N}$ and $G_i=(g_{i,j})_{j=1}^{M_i}$, which preserve the structured geometry of vector objects during multimodal computation.

\subsection{Spatial Perception and Object States}
\label{sec:integration}

Geometry tokens are encoded with spatial organization of the vector scene, while corresponding visual content is distributed across visual tokens $V=(v_p)_{p=1}^{P_v}$. We first enable cross-modal interaction between these tokens inside the VLM and then organize the resulting representations around scene objects. Let
$\mathcal{P}=[V;G;Q]$ denote the visual, geometry, and optional numerical-query tokens. At selected VLM layers, tokens in $\mathcal{P}$ attend bidirectionally to one another, while
language tokens retain the original causal attention pattern. This enables geometry, visual content, and numerical spatial queries to exchange information directly through the VLM backbone attention. We denote the resulting visual and geometry states by $\bar V=(\bar v_p)_{p=1}^{P_v}$ and the geometry representation of parcel $i$ by $\bar G_i=(\bar g_{i,j})_{j=1}^{M_i}$.

We then use one learnable object state $O_i$ for each parcel $i$ to aggregate this multimodal
information. For each visual token $\bar v_p$, let $\rho(i,p)\in\{0,1\}$ indicate whether its image footprint overlaps parcel $i$, and define $\bar V_i=\{\bar v_p\mid\rho(i,p)=1\}$.
Each object state attends only
to itself, the geometry states $\bar G_i$ of its parcel, and the visual
states $\bar V_i$ overlapping that parcel. The exact update is given
in Appendix~\ref{app:GeoVec-Training}.
To further encourage image--vector alignment, we introduce an auxiliary patch-association
objective. The geometry states of parcel $i$ are summarized as
$h_i^{G,\mathrm{assoc}}=\operatorname{Pool}(\bar G_i)$, and their association score with visual
token $\bar v_p$ is:
\begin{equation}
s_{ip}
=
\frac{
\left(W_G\operatorname{LN}(h_i^{G,\mathrm{assoc}})\right)^\top
\left(W_V\operatorname{LN}(\bar v_p)\right)
}{
\sqrt{d_s}
},
\qquad
\mathcal{L}_{\mathrm{assoc}}
=
\operatorname{BCE}_{\mathrm{bal}}
\left(
\{s_{ip}\}_{i,p},
\{\rho(i,p)\}_{i,p}
\right).
\end{equation}

where $W_G$ and $W_V$ project geometry and visual states into a shared $d_s$-dimensional space and $\text{LN}$ denotes layer normalization. The pooled representation is used only for this association objective, while the element-level geometry tokens remain intact and linked to their object states for subsequent reasoning in Section~\ref{sec:reasoning}.

\subsection{Query-Conditioned Spatial States and Language Generation}
\label{sec:reasoning}

The object states $O=\{O_i\}_{i=1}^{N}$ organize the scene by object, while the language query determines which object geometry is relevant to the requested spatial relation. We use role states $A$ to bind query roles to scene objects, and relation states $Z$ to represent spatial relations over the corresponding geometries. For each query role $r$, the instruction representation $\bar t$ and a role embedding $e_r$ define a selection over the object states, $\pi_i^{(r)}=\operatorname{softmax}_{i}\left((W_q(\bar t+e_r))^\top W_k O_i/\sqrt{d}\right)$. This language-conditioned binding is supervised by $\mathcal{L}_{\mathrm{select}}$. The bound object determines its linked element-level geometry memory $H_r$.
$\mathcal{F}_{A}$ aggregates this geometry into a role state $A_r=\mathcal{F}_{A}\!\left(\pi^{(r)},H_r\right)$, allowing each query role to retain access to its associated element-level geometry. For a spatial instance $e$ involving roles $R_e$, the corresponding role states and geometry memories are further composed by $\mathcal{F}_{Z}$ into a relation state, $Z_e=\mathcal{F}_{Z}\!\left(\{A_r,H_r\}_{r\in R_e}\right)$. Here, $O$ organizes vector and visual content by object, $A$ connects language-conditioned roles to the corresponding geometry, and $Z$ represents relation-specific spatial evidence over the selected geometries. These states are updated at selected reasoning layers of the VLM.

The resulting role and relation states form the spatial memory
$M_{\mathrm{sp}}=[A;Z]$, which is read by the language states through gated cross-attention:
\begin{equation}
\widetilde L
=
L
+
\tanh(\alpha)\,
\operatorname{CrossAttn}_{\mathrm{VLM}}
\left(
L,\,
M_{\mathrm{sp}},\,
M_{\mathrm{sp}}
\right).
\end{equation}
Structured decoders attached to the corresponding spatial states provide auxiliary supervision during training, while their predictions are not used by the language path. The overall objective is
\begin{equation}
\mathcal{L}
=
\mathcal{L}_{\mathrm{LM}}
+
\lambda_{\mathrm{assoc}}\mathcal{L}_{\mathrm{assoc}}
+
\lambda_{\mathrm{select}}\mathcal{L}_{\mathrm{select}}
+
\sum_{k\in\mathcal K}\lambda_k\mathcal{L}_k ,
\end{equation}
where $\mathcal{L}_k$ denotes a structured objective applied when the corresponding supervision is available. The stage-wise training procedure is described in Appendix~\ref{app:GeoVec-Training}.

\section{Experimental Setup}

\paragraph{Dataset.}
We construct a geospatial reasoning dataset from real street-block polygons, where each scene pairs a \(512\times512\) raster image with spatially aligned vector geometry. The China dataset spans 221 city-level regions and contains 63,973 scenes, with 359,340 question–answer instances used for training. We split the data geographically into disjoint training, validation, and test regions, and restrict geometry-encoder pretraining to the training regions. For zero-shot geographic transfer, we further evaluate on six U.S. metropolitan regions (900 scenes) and Singapore (300 scenes), neither of which contributes to training or model selection. Full construction procedures and split statistics are provided in Appendix~\ref{app:dataset}.

\paragraph{Tasks and Evaluation.}
We evaluate five spatial reasoning categories covering six task families, including eight-way direction, comparative proximity, parcel–parcel topology, point–parcel topology, area selection, and point to parcel boundary distance selection. Direction additionally includes continuous horizontal and vertical displacement ($\Delta x,\Delta y$). The China test set contains 1,800 questions per task, and the same protocol is applied to the U.S. and Singapore transfer sets, with 1,800 and 600 questions per task, respectively. To reduce shortcut solutions, we exclude ambiguous cases and include hard samples where centroid distance, bounding-box area, or nearest-vertex distance disagrees with the exact geometric relation. Reversible queries are evaluated in pairs for consistency. We report accuracy as the primary metric, together with reciprocal consistency and component-wise RMSE where applicable. Full construction and evaluation details are provided in Appendix~\ref{app:task-construction}.

\paragraph{Baselines and Input Interfaces.}
We evaluate GeoVec-VLM from two complementary perspectives. First, we compare against general-purpose and spatial VLMs using the geometry interfaces available to them. Qwen3-VL-8B serves as the primary same-backbone reference under image-only and image-plus-GeoJSON inputs. GeoJSON is a widely used format for encoding vector geographic data. Larger general-purpose VLMs are evaluated under the same supported interfaces as external capability references, while SpatialRGPT \citep{cheng2024spatialrgpt} and SR-3D \citep{cheng2026sr3d} are fine-tuned on our dataset with ground-truth region masks. Second, to isolate the effect of geometry representation, we conduct controlled comparisons under the same Qwen3-VL-8B backbone, training data, and task supervision using serialized coordinates, Geo2Vec \citep{chu2026geo2vec}, Poly2Vec \citep{siampou2025poly2vec}, a pooled GeoVec representation, and our element-level GeoVec interface. These controlled variants distinguish the benefit of providing vector information from the benefit of preserving its fine-grained geometric structure during multimodal reasoning.

\begin{table*}[h]
\centering
\caption{
Spatial reasoning on geographically disjoint test regions from China.
$^\dagger$Area and Distance are evaluated through language generation without corresponding structured decoders or downstream task supervision.
SpatialRGPT$^{*}$ \citep{cheng2024spatialrgpt} and SR-3D$^{*}$ \citep{cheng2026sr3d}
are fine-tuned from released checkpoints on our dataset. Masks denote ground-truth region masks.
}
\label{tab:main-results}

\vspace{4pt}

\small
\setlength{\tabcolsep}{3pt}
\renewcommand{\arraystretch}{1.03}

\newcolumntype{Y}{>{\centering\arraybackslash}X}

\begin{tabularx}{\textwidth}{
    >{\raggedright\arraybackslash}p{0.19\textwidth}
    >{\raggedright\arraybackslash}p{0.15\textwidth}
    YYYYY
}
\specialrule{1.1pt}{0pt}{0pt}

Model
& Interface
& Direction $\uparrow$
& Proximity $\uparrow$
& Topology $\uparrow$
& Area$^\dagger$ $\uparrow$
& Distance$^\dagger$ $\uparrow$ \\

\midrule

\rowcolor{gray!15}
Qwen3-VL-8B
& image
& 0.312 & 0.553 & 0.539 & 0.528 & 0.286 \\

\rowcolor{gray!15}
& + GeoJSON
& 0.322 & 0.566 & 0.508 & 0.602 & 0.336 \\

DeepSeek-V4-Flash
& image
& 0.635 & 0.562 & 0.548 & 0.739 & 0.417 \\

& + GeoJSON
& 0.609 & 0.569 & 0.631 & 0.744 & 0.475 \\

\rowcolor{gray!15}
Kimi K2.5
& image
& 0.668 & 0.714 & 0.652 & 0.756 & 0.591 \\

\rowcolor{gray!15}
& + GeoJSON
& 0.655 & 0.711 & 0.693 & 0.711 & 0.605 \\

GPT-5.1
& image
& 0.614 & 0.733 & 0.651 & 0.691 & 0.586 \\

& + GeoJSON
& 0.581 & \underline{0.763} & \underline{0.711} & 0.716 & 0.621 \\

\rowcolor{gray!15}
Gemini 2.5 Flash
& image
& 0.640 & 0.636 & 0.584 & \textbf{0.784} & 0.622 \\

\rowcolor{gray!15}
& + GeoJSON
& 0.629 & 0.633 & 0.614 & 0.766 & 0.626 \\

Qwen 3.7 Plus
& image
& 0.667 & 0.706 & 0.635 & 0.748 & \underline{0.643} \\

& + GeoJSON
& \underline{0.683} & 0.716 & 0.707 & 0.749 & 0.621 \\

\midrule

\rowcolor{gray!15}
SpatialRGPT$^{*}$
& image + masks
& 0.436 & 0.644 & 0.597 & 0.513 & 0.351 \\

\rowcolor{gray!15}
SR-3D$^{*}$
& image + masks
& 0.325 & 0.497 & 0.570 & 0.593 & 0.298 \\

\midrule

\textbf{GeoVec-VLM-8B}
& \textbf{Native vector}
& \textbf{0.804}
& \textbf{0.807}
& \textbf{0.811}
& \underline{0.778}
& \textbf{0.659} \\

\specialrule{1.1pt}{0pt}{0pt}
\end{tabularx}

\end{table*}

\section{Results and Analysis}

\subsection{Spatial Reasoning Accuracy and Geometric Coherence.}
\label{sec:main-results}

Table~\ref{tab:main-results} shows that general-purpose VLMs can recover some spatial relations from images, but adding vector geometry expressed as GeoJSON text does not consistently improve performance and can even reduce accuracy, as seen for Kimi K2.5 on Direction and Gemini 2.5 Flash on Area. GeoVec-VLM consistently improves the three supervised tasks, exceeding
the strongest image/GeoJSON baseline by 4.4--12.1 percentage points. These gains extend beyond directly supervised objectives: GeoVec-VLM remains competitive
on Area and achieves the best Distance accuracy, even though these were excluded from structured decoders. In our adaptation setting, fine-tuned SpatialRGPT and SR-3D perform comparably to Qwen3-VL-8B, suggesting limited benefit from image- and region-mask-based spatial specialization for metric reasoning over explicit parcel geometry. GeoVec-VLM maintains performance under zero-shot geographic transfer, achieving 77.5\% mean accuracy in the U.S. and 78.3\% in Singapore,
compared with 77.2\% on the China test set
(Appendix~\ref{app:Geographic Transfer}).

\begin{figure}[h]
\centering
\includegraphics[width=\linewidth]{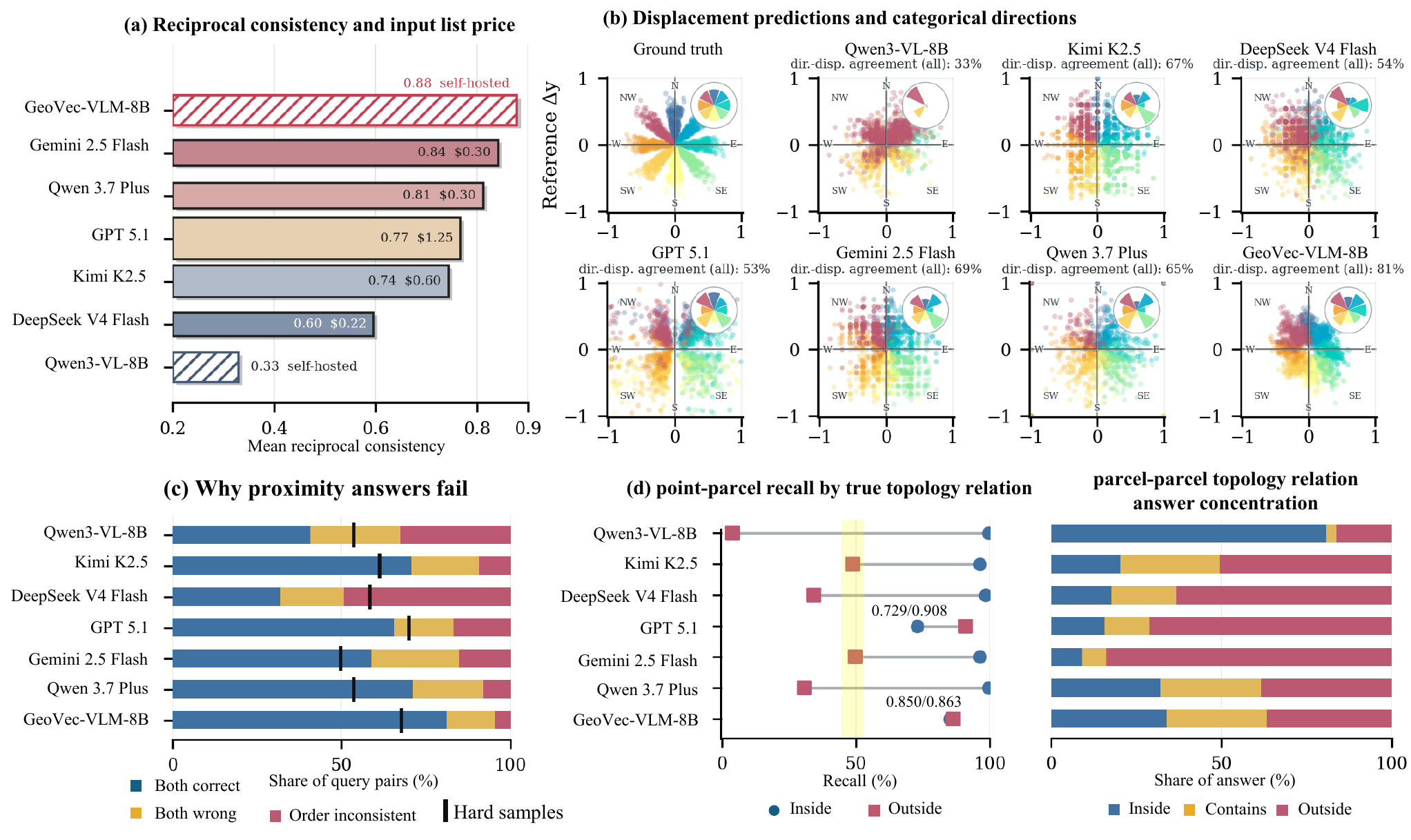}
\setlength{\abovecaptionskip}{0pt}
\caption{\textbf{Task accuracy can conceal geometrically inconsistent spatial behavior.}
(a) Mean reciprocal consistency over direction, comparative proximity, and topology. Because repeated errors can remain reciprocally consistent, panels (b--d) examine task-specific geometric constraints. (b) Categorical direction predictions are compared with the directions implied by predicted horizontal and vertical displacements. Polar histograms show the corresponding answer distributions. (c) Proximity is evaluated under candidate-order reversal, which preserves the physically closer parcel. Black ticks indicate hard-sample accuracy. (d) Point--parcel class-conditional recall and parcel--parcel prediction frequencies reveal query-dependent topology bias.}
\label{fig:consistency}
\end{figure}

Task accuracy alone is insufficient to assess whether a model captures the
underlying spatial relations. As shown in Fig.~\ref{fig:consistency}(b),
baseline outputs can change even when the geometry is unchanged. GPT-5.1, for
example, achieves only 53\% agreement between categorical direction and the
direction implied by its displacement estimate. Proximity decisions can also
flip after candidate-order reversal, while topology outputs show a strong bias
towards \texttt{inside} or \texttt{outside}. GeoVec-VLM is much less sensitive
to these effects, reaching 81\% agreement between direction and displacement
and the highest reciprocal consistency of 0.88, while producing more stable and
balanced results on proximity and topology. We attribute this stability to the
explicit geometric structure preserved by vector data, which provides a direct
basis for spatial reasoning.

\subsection{ABLATION AND MECHANISM ANALYSIS}
\label{sec:ablations}

To understand where GeoVec-VLM’s gains come from, we analyze three parts of the model. Different geometry representations are first compared to validate whether preserving fine-grained structure is necessary. Cross-modal grounding and perturbation tests then assess whether vector objects can be reliably aligned with their corresponding visual regions. Finally, ablations on the spatial states and their supervision show how the encoded geometry is used in language generation. 

\begin{table}[H]
\vspace{-8pt}
\centering
\caption{\textbf{Retaining structured vector representations through multimodal interaction improves spatial reasoning.}
All variants use Qwen3-VL-8B with the same training split and task setting. Geo2Vec$^{*}$ and Poly2Vec$^{*}$ are our VLM adaptations of the open-source geometry encoders under this controlled setting. Pool GeoVec-VLM pools projected geometry tokens to a single embedding.}
\label{tab:representation}
\vspace{4pt}
\small
\setlength{\tabcolsep}{3pt}

\begin{tabular*}{\textwidth}{@{\extracolsep{\fill}}lccccc@{}}
\specialrule{1.2pt}{0pt}{0pt}
Geometry Representation
& Direction $\uparrow$
& Proximity $\uparrow$
& Topology $\uparrow$
& Area $^\dagger$ $\uparrow$
& Distance $^\dagger$ $\uparrow$ \\
\midrule

Geo2Vec$^{*}$ \citep{chu2026geo2vec}
& 0.203 & 0.511 & 0.526 & 0.567 & 0.246 \\

Poly2Vec$^{*}$ \citep{siampou2025poly2vec}
& 0.233 & 0.503 & 0.525 & 0.566 & 0.248 \\

Serialized GeoJSON
& 0.249 & 0.517 & 0.569 & \underline{0.613} & 0.332 \\

Pool GeoVec-VLM
& \underline{0.704} & \underline{0.756} & \underline{0.748} & 0.603 & \underline{0.568} \\

\textbf{GeoVec-VLM}
& \textbf{0.804}
& \textbf{0.807}
& \textbf{0.811}
& \textbf{0.778}
& \textbf{0.659} \\

\specialrule{1.2pt}{0pt}{0pt}
\label{tab: geometry inferace}
\end{tabular*}

\end{table}

\paragraph{Structure-preserving representation matters}

The form in which vector geometry is exposed to the VLM substantially affects spatial reasoning. Table~\ref{tab:representation} compares several geometry interfaces under the
same backbone and training setup. Geo2Vec and Poly2Vec, despite encoding geometric properties of complete spatial objects, perform close to or below serialized GeoJSON on several tasks, indicating that object-level geometry encoding alone may not retain enough structure for spatial reasoning. In contrast, Pool GeoVec-VLM already provides gains, and retaining the point- and path-level states throughout multimodal interaction improves all five tasks by a further 5.1--17.5 points. Since the pooled and full variants share the same geometry encoder and training setup, this difference isolates the benefit of preserving
element-level geometric structure during reasoning. We further examine geometry-token components in Appendix~\ref{app:geometry_intervention}, and report input-token consumption together with task accuracy in Appendix~\ref{app:input token consumption}.

\begin{figure}[h]
\centering
\includegraphics[width=\linewidth]{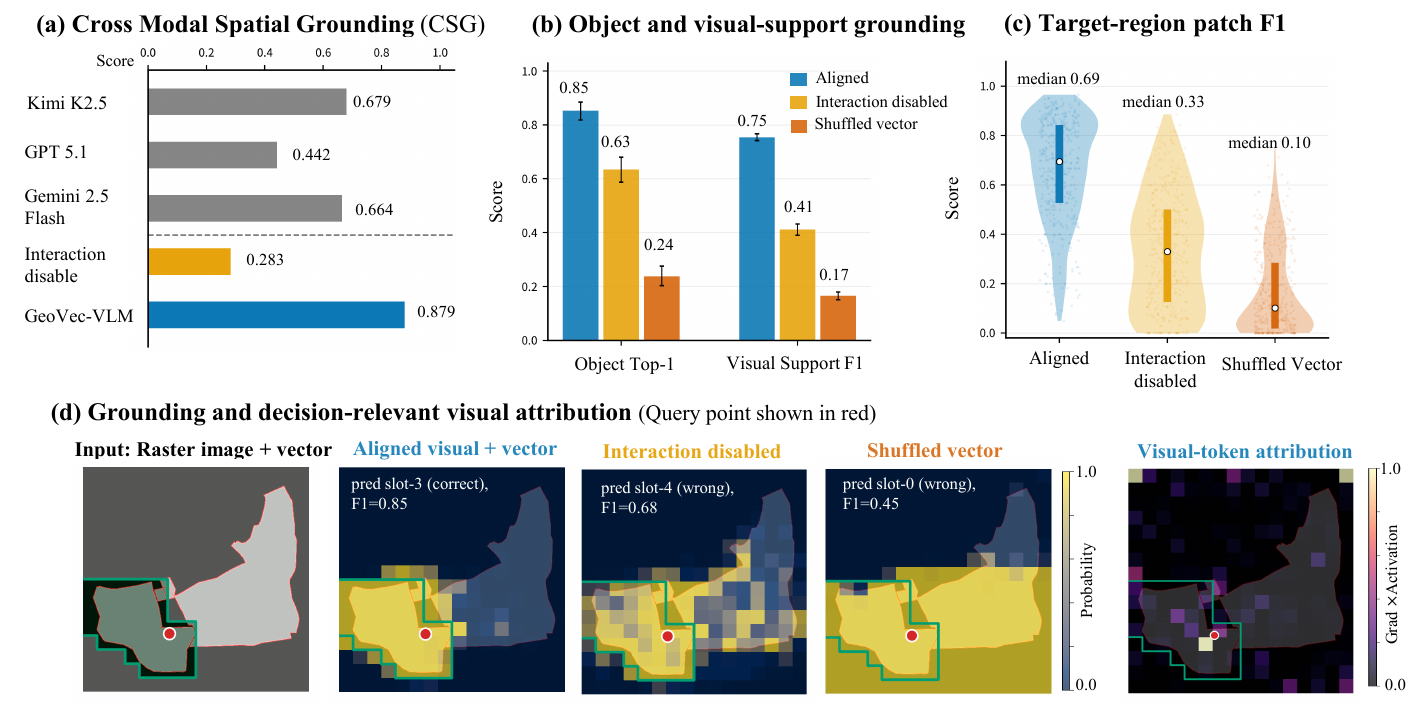}
\setlength{\abovecaptionskip}{0pt}
\caption{\textbf{Vector–visual alignment supports cross-modal grounding.}
(a) CSG requires mapping image-only query points to their corresponding vector parcels. (b–c) Grounding signals weaken when vector–visual interaction is disabled, and further deteriorate after shuffling vector geometry. (d) With aligned inputs, visual support and decision-relevant attribution focus on the corresponding parcel. The final panel shows positive Gradient$\times$Activation for the gold grounding margin.}
\label{fig:alignment}
\end{figure}

\paragraph{Vector–visual alignment}
Given that the query point is visible only in the image, can the model identify its corresponding parcel from vector data? As shown in Fig.~\ref{fig:alignment}(a), GeoVec-VLM achieves 87.9\% accuracy on this cross-modal grounding task, outperforming Kimi K2.5 by 20.0 percentage points. This capability depends strongly on vector–visual interaction, as disabling the interaction reduces accuracy to 28.3\%. Fig.~\ref{fig:alignment}(b-c) further show that, with aligned inputs, the model more reliably selects the correct parcel and concentrates visual support within the corresponding region, reaching 0.85 Object Top-1 and 0.75 Visual Support F1. When the vector geometry is shuffled, these scores drop to 0.24 and 0.17, respectively, while the median target-region patch F1 decreases from 0.69 to 0.10. The example in Fig.~\ref{fig:alignment}(d) makes this effect spatially explicit, and Gradient×Activation attribution further localizes the grounding decision to visual tokens around the corresponding region.

\begin{figure}[h]
\centering

% =========================================================
% Main contents
% =========================================================
\begin{minipage}[t]{0.64\linewidth}
\vspace{0pt}
\centering

\small
\setlength{\tabcolsep}{2.8pt}
\renewcommand{\arraystretch}{1.25}

\begin{tabular*}{\linewidth}
{@{\extracolsep{\fill}}lccccc@{}}
\specialrule{1.2pt}{0pt}{0pt}
Variant
& Dir. $\uparrow$
& Prox. $\uparrow$
& Topo. $\uparrow$
& Area $\uparrow$
& Dist. $\uparrow$ \\
\midrule

\textbf{Full}
& \textbf{0.804}
& \textbf{0.807}
& \textbf{0.811}
& \textbf{0.778}
& \textbf{0.659} \\

w/o $O/A/Z$
& 0.453
& 0.507
& 0.573
& 0.372
& 0.460 \\

w/o Struct. Dec.
& 0.747
& 0.632
& 0.777
& 0.499
& 0.406 \\

w/o R--O Bind.
& 0.796
& 0.772
& 0.807
& 0.746
& 0.622 \\

\specialrule{1.2pt}{0pt}{0pt}
\end{tabular*}\par
\vskip 10pt

\refstepcounter{table}
\label{tab:ablation}
{\footnotesize
\justifying
Table~\thetable: \textbf{Ablation of spatial states and supervision}.
All values report final-answer accuracy. The w/o $O/A/Z$ variant retains the geometry tokens $G$ but removes the learned spatial states. The w/o Struct. Dec. variant removes the auxiliary structured decoders, whereas the w/o R–O Bind. removes language-conditioned role--object binding.
\par}

\end{minipage}%
\hfill%
\begin{minipage}[t]{0.32\linewidth}
\vspace{0pt}
\centering

\includegraphics[
    width=\linewidth
]{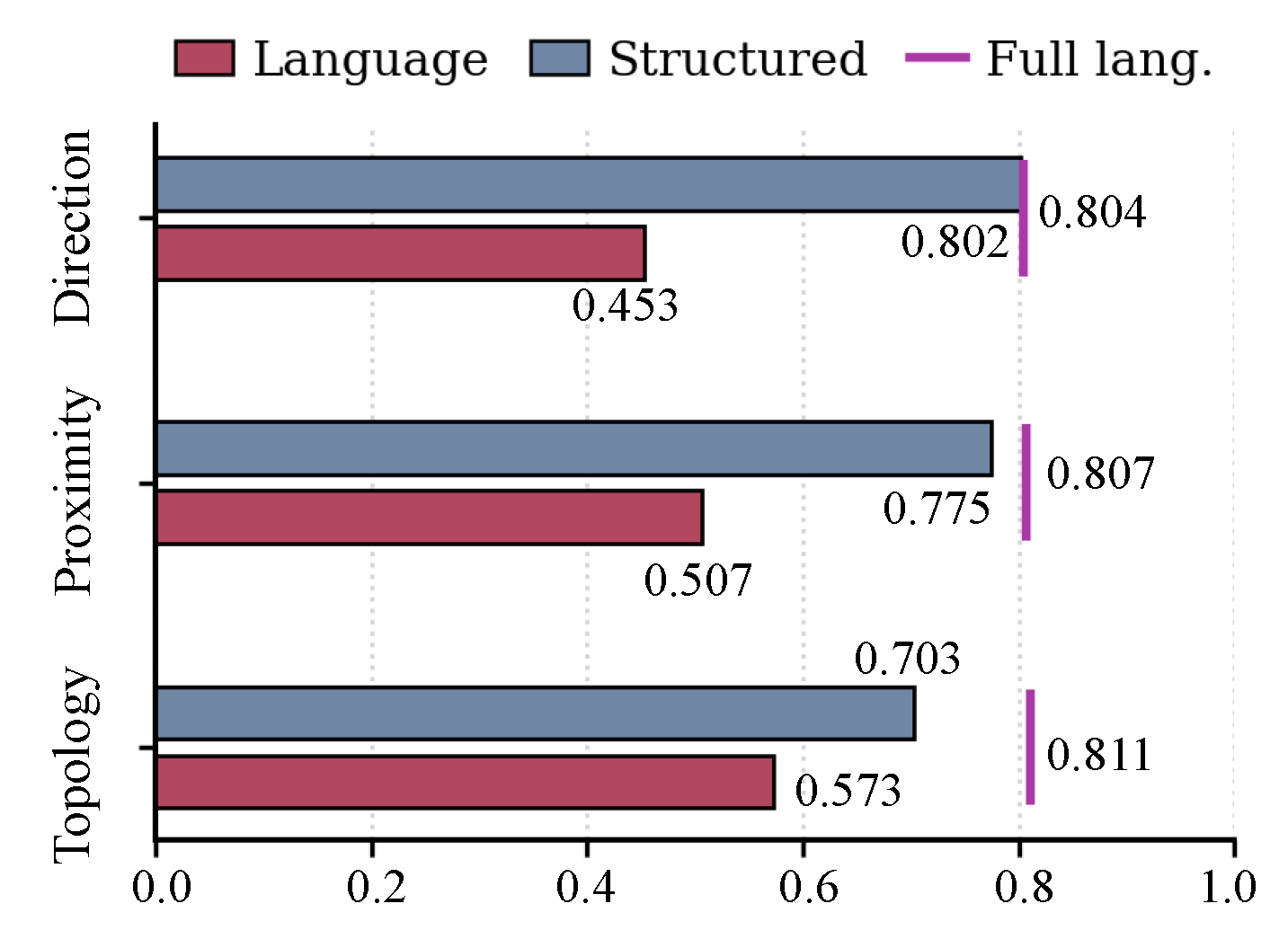}

\vspace{2pt}

\refstepcounter{figure}
\label{fig:ablation}
{\footnotesize
\justifying
Figure~\thefigure: Language and structured readouts without $O/A/Z$, attached directly to $G$.
\par}

\end{minipage}

\end{figure}

\paragraph{Spatial states in language generation}
Table~\ref{tab:ablation} examines what is required for encoded geometry to support multimodal reasoning. Removing $O/A/Z$ reduces answer accuracy by 19.9--40.6 percentage points across the five tasks, while the structured readouts remain high at 70.3 to 80.2\% in Fig.~\ref{fig:ablation}. This gap indicates that $G$ retains substantial spatial information, whereas producing correct answers requires the $O/A/Z$ states to organize geometry together with visual information and language-conditioned query roles. Structured supervision further improves answer accuracy by 3.4 to 27.9 points. Notably, Area and Distance improve by 27.9 and 25.3 points despite having no corresponding structured decoders, suggesting that spatial supervision transfers beyond the directly supervised relations. Role--object binding has smaller effects when candidate roles are explicit (0.4--3.7 points), but yields about a 5-point gain when the referred object must be resolved from language, consistent with its role in selecting the relevant object geometry. Further analysis in Appendix~\ref{app:language_spatial_fusion} traces how these spatial states affect the language path.

\section{Conclusion}

Spatial reasoning requires reasoning over the geometry that relates objects in a scene. In this work, we explore geospatial vector geometry as a native spatial modality for VLMs. We find that organizing multimodal reasoning around this geometry provides an effective spatial substrate, improving both spatial reasoning accuracy and geometric consistency. We note that accurate answers do not necessarily imply geometrically consistent reasoning. A model may answer individual questions correctly while its predictions across related questions fail to agree with the same underlying geometry. We also find that geometric information can remain recoverable inside a VLM without being effectively used by its language predictions. We hypothesize that robust spatial reasoning emerges when different predictions are grounded in a shared internal representation of the underlying geometry. We hope this perspective will inspire further work on spatial structure as a native component of multimodal models. \textbf{Limitations}: we study settings where accurate vector geometry and raster imagery are both available. This lets us focus on how spatial structure can be represented and used inside a VLM, while leaving noisy, incomplete, outdated, or misaligned vector data for future work. Our experiments are limited to static 2D polygonal scenes. Other geometry types and broader visual settings remain to be explored. Comparisons with proprietary and released spatial VLMs are used as capability references rather than controlled architectural comparisons.

\bibliography{references}
\bibliographystyle{iclr2027_conference}

\clearpage
\appendix

% =========================================================
\section{Additional Method and Training Details}
\label{app:method}

% 补充正文 Method 中省略的实现细节。
% 包括 geometry element construction、geometry encoder、
% spatial states、structured objectives、language--spatial fusion
% 以及 stage-wise training procedure。

\subsection{Vector Geometry Representation}
\label{app:geometry}

This section provides the implementation details of the vector encoder introduced in Section~\ref{sec:geometry-encoding}. We tokenize each parcel as a sequence of reference-point and path elements. The encoder first models the geometry within each parcel polygon and then jointly models the relationships among all parcel elements in a scene.

% =========================================================
\paragraph{Geometry tokenization and element features.}
Our tokenization follows two structural principles from established
geospatial representations that explicitly
preserve spatial structure in geospatial tokenization~\citep{huang2025spok}. 
Polygon geometry in GIS is represented by
ordered rings with explicit exterior and interior boundaries; accordingly,
we preserve part and ring membership and ring orientation
following GeoJSON conventions~\citep{butler2016geojson}. OpenLR represents
a path using an ordered sequence of reference points together with
attributes that retain geometric information between successive
references~\citep{tomtomdlr2012openlr}. We adapt this reference-point/path
decomposition to polygon boundaries: retained boundary vertices form
reference-point elements, while each interval along the original ring is
represented by a path element carrying the geometry omitted by
simplification. This representation reduces the explicit boundary sequence
while preserving boundary order, connectivity, and local metric structure.
All geometries are projected into a region-specific metric coordinate
reference system before these features are computed.

\begin{figure}[H]
    \centering
    \includegraphics[width=1\linewidth]{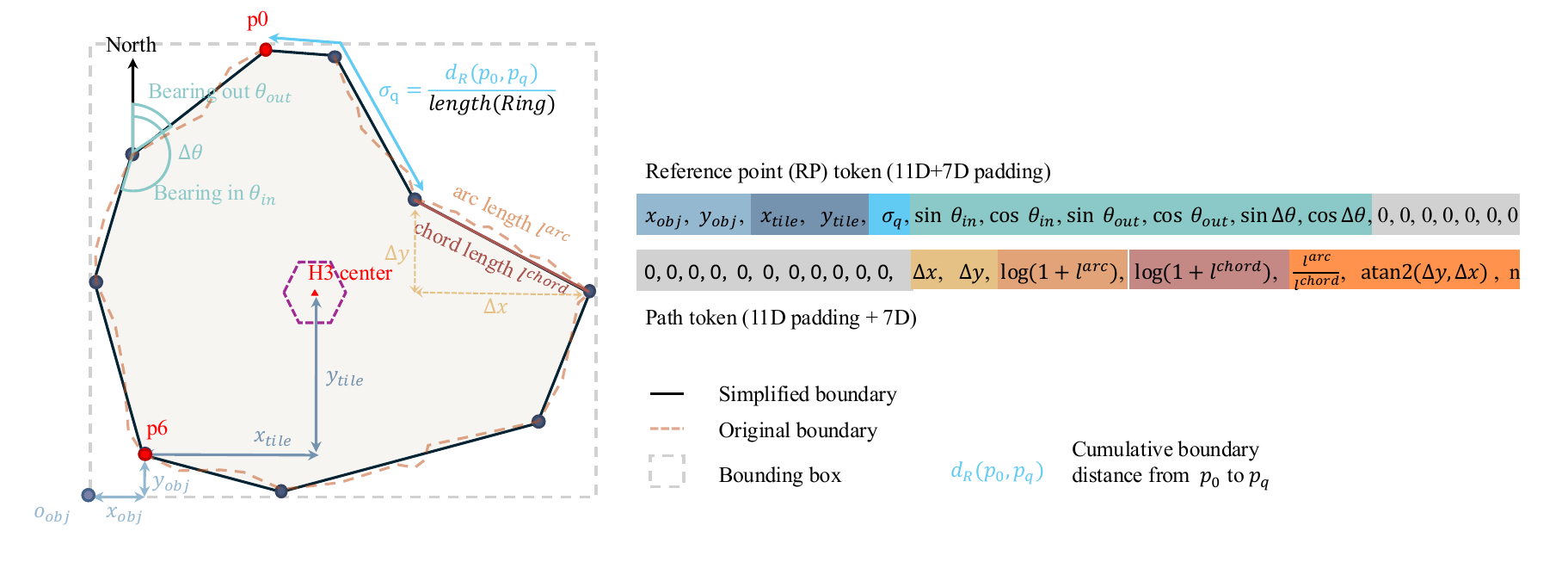}
    \caption{Reference-point and path feature construction. 
    \textbf{Left}: boundary simplification and the geometric features extracted for reference-point and path elements. \textbf{Right}: the features are packed into a shared 18-dimensional feature representation, where reference-point and path features occupy complementary channels and inactive channels are zero-padded. The complete feature definitions  are summarized in Table~\ref{tab:geometry-features}.
    }
    \label{fig:geometry-features}
\end{figure}

We apply Douglas--Peucker simplification independently to each ring according to the complexity, with a maximum tolerance of 2 meters. The retained vertices form reference points, and each pair of consecutive reference points defines the path element. For a ring $R$ with simplified reference points $\{p_j\}_{j=0}^{m-1}$, let $\gamma_j$ denote the path from $p_j$ to $p_{(j+1)\bmod m}$. We serialize the ring as 

\begin{equation}
X_R =
\left[
x^{\mathrm{rp}}_0,
x^{\mathrm{path}}_0,
x^{\mathrm{rp}}_1,
x^{\mathrm{path}}_1,
\ldots,
x^{\mathrm{rp}}_{m-1},
x^{\mathrm{path}}_{m-1}
\right].
\label{eq:parcel-serialization}
\end{equation}

The ring sequences belonging to the same parcel are concatenated to form the object sequence $X_i$. We record the part and ring indices to preserve  the structural membership of each element. We extract complementary geometry- and topology-related features from reference-point and path elements using a shared 18-dimensional interface. As shown in Figure~\ref{fig:geometry-features}, reference-point elements occupy the first 11 channels and path elements use the remaining seven channels, with inactive channels zero-padded. We denote the element feature vector by $f_q$.

\begin{table}[H]
\centering
\caption{
Geometric features of reference-point and path elements.}
\label{tab:geometry-features}

\small
\setlength{\tabcolsep}{3.2pt}
\renewcommand{\arraystretch}{1.12}

\begin{tabular}{@{}p{0.21\linewidth}p{0.25\linewidth}
                    p{0.20\linewidth}p{0.26\linewidth}@{}}
\toprule
\textbf{RP feature}
& \textbf{Geometric interpretation}
& \textbf{Path feature}
& \textbf{Geometric interpretation} \\
\midrule

$(x_{\mathrm{obj}},y_{\mathrm{obj}})$
& Position normalized by the bounding box
& $(\Delta x,\Delta y)/1000$
& Path endpoint displacement \\

$(x_{\mathrm{tile}},y_{\mathrm{tile}})/1000$
& Offset from the parcel's H3-cell center
& $\log(1+l^{\mathrm{arc}})$
& Length of the original boundary path \\

$\sigma_q =
\frac{d_R(p_0,p_q)}
     {\operatorname{length}(Ring)}$
& Normalized position along the ring
& $\log(1+l^{\mathrm{chord}})$
& Chord length between path endpoints \\x

$(\sin\theta_{\mathrm{in}},\cos\theta_{\mathrm{in}})$
& Incoming direction
& $
\frac{l^{\mathrm{arc}}}
     {l^{\mathrm{chord}}}$
& How curvy compared to the straight chord  \\

$(\sin\theta_{\mathrm{out}},\cos\theta_{\mathrm{out}})$
& Outgoing direction
& $\operatorname{atan2}(\Delta y,\Delta x)$
& Chord bearing \\

$(\sin\Delta\theta,\cos\Delta\theta)$
& Signed local turning angle
& $n$
& Number of intermediate original vertices removed by simplification \\

\bottomrule
\end{tabular}
\end{table}

Table~\ref{tab:geometry-features} summarizes the features and their geometric interpretation. These features are designed to preserve the spatial information most relevant to boundary reasoning: reference-point features encode location, boundary progression, and local turning geometry, while path features retain metric and shape information from the original boundary intervals that would otherwise be lost during simplification. Together, they reduce the number of explicit boundary elements while preserving both local shape and metric structure.

% =========================================================
\subsection{Vector Geometry Encoder Pretraining}

\begin{figure}[H]
    \centering
    \includegraphics[width=\linewidth]{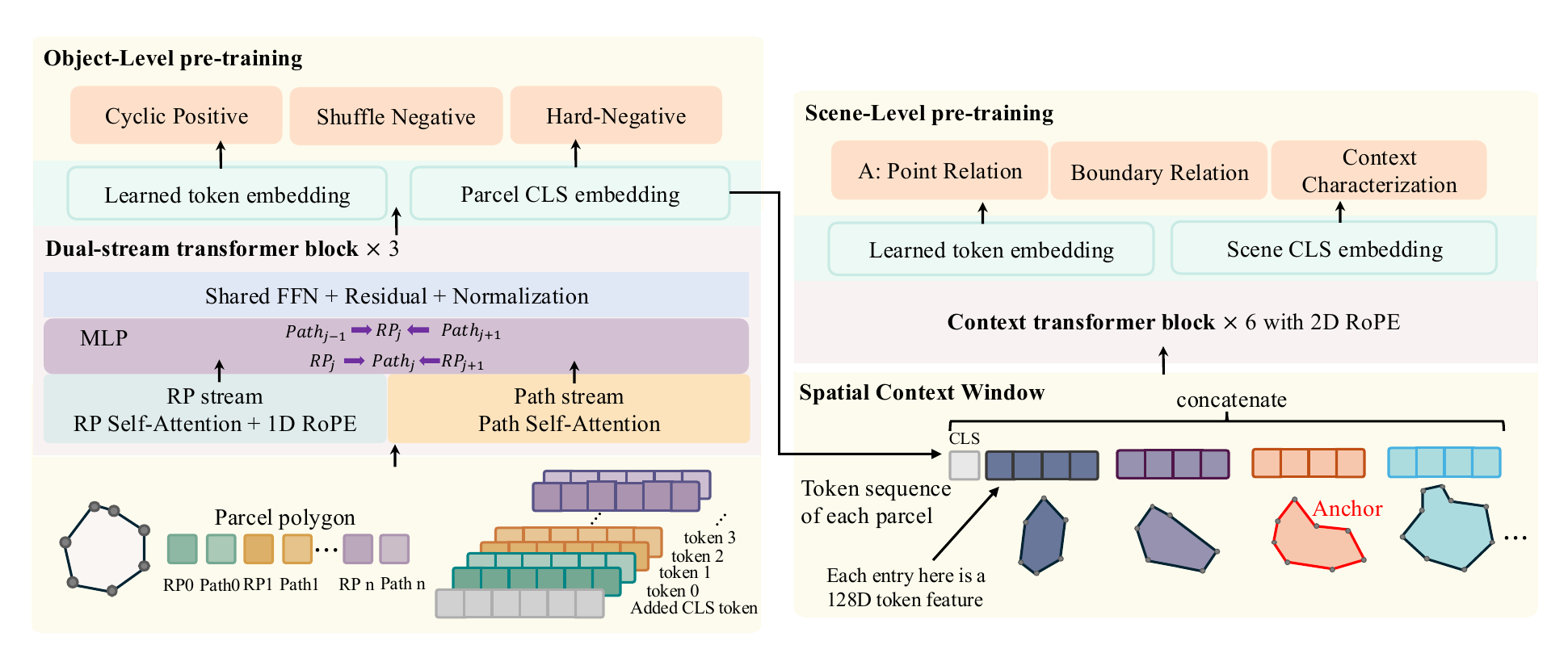}
    \caption{
    \textbf{Vector encoder architecture and geometry pretraining.}
    The object encoder processes alternating reference-point and path elements with separate attention streams and endpoint-conditioned cross-stream interaction. The resulting element sequences are then concatenated into a spatial context window and processed by the scene encoder with 2D positional encoding. }
    \label{fig:geometry-pretraining}
\end{figure}

We pretrain the vector geometry encoder in two stages. Object-level pretraining learns the boundary structure within each polygon, and scene-level pretraining subsequently models spatial relations across neighboring polygons. Figure~\ref{fig:geometry-pretraining} summarizes the overall pipeline.

\paragraph{Object-level pretraining.}

After feature standardization using means and standard deviations computed from the training split, each 18-D feature $\widehat f_q$ is then linearly projected into the geometry hidden space. We additionally encode polygon membership and token type with learned embeddings to the same 128-dimension spaces and add them to the projected feature representation. 

As illustrated on the left of Figure~\ref{fig:geometry-pretraining}, the object encoder consists of three dual-stream Transformer blocks with four attention heads. Reference-point and path elements are processed by separate attention streams. Reference-point attention applies one-dimensional RoPE using the normalized arc position $\sigma_q$, while cross-stream MLPs exchange information between incident reference points and paths. The encoder produces contextualized element states $\widetilde U_i=(\tilde u_{i,j})_{j=1}^{M_i}$, and a learned CLS query performs attention pooling over these states to obtain the parcel representation.

Object-level pretraining targets three complementary properties of polygon boundaries: invariance to the arbitrary starting point, sensitivity to boundary order, and discrimination between embedding-similar parcels. To learn cyclic invariance, we use a symmetric NT-Xent loss $\mathcal L_{\mathrm{cyclic}}$ that treats a one-pair cyclic shift as the positive view and other parcels in the batch as negatives ($\tau_{\mathrm{cyclic}}=0.1$). Both views are perturbed with Gaussian feature noise ($\sigma=0.01$) and token feature-vector dropout ($p=0.05$). To preserve boundary order, a cosine margin-ranking loss $\mathcal L_{\mathrm{order}}$ requires the cyclic view to be more similar to the original than a randomly permuted view ($m=0.1$). Finally, a hard-negative InfoNCE loss $\mathcal L_{\mathrm{hard}}$ improves discrimination using $K=5$ embedding-similar parcels ($\tau_{\mathrm{hard}}=0.05$), mined offline after an initial round of pretraining.

Let $h_i$, $h_i^{+}$, and $h_i^{-}$ denote the L2-normalized parcel representations of the original, cyclic, and shuffled views, respectively, and let $h_{i,k}^{h}$ denote the $k$-th hard negative of parcel $i$. Using cosine similarity $\operatorname{sim}(\cdot,\cdot)$, we define
\begin{equation}
s_{ij}^{+}
=
\operatorname{sim}(h_i,h_j^{+}),
\qquad
s_i^{-}
=
\operatorname{sim}(h_i,h_i^{-}),
\qquad
s_{ik}^{h}
=
\operatorname{sim}(h_i,h_{i,k}^{h}).
\end{equation}

\begin{equation}
\mathcal L_{\mathrm{cyclic}}
=
-\frac{1}{2B}
\sum_{i=1}^{B}
\left[
\log
\frac{
\exp\!\left(s_{ii}^{+}/\tau_{\mathrm{cyclic}}\right)
}{
\sum_{j=1}^{B}
\exp\!\left(s_{ij}^{+}/\tau_{\mathrm{cyclic}}\right)
}
+
\log
\frac{
\exp\!\left(s_{ii}^{+}/\tau_{\mathrm{cyclic}}\right)
}{
\sum_{j=1}^{B}
\exp\!\left(s_{ji}^{+}/\tau_{\mathrm{cyclic}}\right)
}
\right].
\label{eq:cyclic-loss}
\end{equation}

\begin{equation}
\mathcal L_{\mathrm{order}}
=
\frac{1}{B}
\sum_{i=1}^{B}
\max
\left(
0,\,
m-s_{ii}^{+}+s_i^{-}
\right).
\label{eq:order-loss}
\end{equation}

\begin{equation}
\mathcal L_{\mathrm{hard}}
=
-\frac{1}{B}
\sum_{i=1}^{B}
\log
\frac{
\exp\!\left(s_{ii}^{+}/\tau_{\mathrm{hard}}\right)
}{
\exp\!\left(s_{ii}^{+}/\tau_{\mathrm{hard}}\right)
+
\displaystyle\sum_{k=1}^{K}
\exp\!\left(s_{ik}^{h}/\tau_{\mathrm{hard}}\right)
}.
\label{eq:hard-loss}
\end{equation}

Here,
$\tau_{\mathrm{cyclic}}=0.1$,
$\tau_{\mathrm{hard}}=0.05$,
$m=0.1$, and $K=5$.
The complete object-level objective is
\begin{equation}
\mathcal L_{\mathrm{obj}}
=
\mathcal L_{\mathrm{cyclic}}
+
5\,\mathcal L_{\mathrm{order}}
+
\mathcal L_{\mathrm{hard}}.
\label{eq:object-pretraining-loss}
\end{equation}

\paragraph{Scene-level pretraining.}

To model cross-parcel spatial relations, we assemble object-encoder element states into an anchor-centered local scene. For each parcel, a 4-D scene-relative feature $r_i=[\Delta I_i,\Delta J_i,\delta x_i^{\mathrm{cell}}, \delta y_i^{\mathrm{cell}}]$, combining H3-cell displacement and within-cell centroid offset, is projected and added to its element states together with a learned parcel-slot embedding. Anchor-relative element coordinates are supplied to 2D RoPE for fine-grained spatial encoding. Reference points use their own coordinates and paths use endpoint midpoints. The concatenated sequence is then processed by the scene encoder, yielding contextualized element states $U=\{U_i\}_{i=1}^{N}$ and a scene-level CLS representation.

As illustrated on the right of Figure~\ref{fig:geometry-pretraining}, we pretrain the scene encoder with three spatial objectives. The \emph{point-relation} task predicts the containing parcel (or outside) and the distance to the nearest parcel boundary. The \emph{boundary-relation} task predicts a four-bin gap category and the exact distance between a queried pair of boundary segments. For \emph{context characterization}, a regression head operating on the scene CLS representation predicts six scene-level statistics: log bounding-box area, parcel density, the mean and standard deviation of log parcel area, coverage ratio, and fragmentation. The corresponding objectives are:
\begin{align}
\mathcal L_{\mathrm{point}}
&=
\operatorname{CE}(\hat y_p,y_p)
+
0.25\,
\operatorname{SmoothL1}(\hat d_p,\tilde d_p),
\\
\mathcal L_{\mathrm{boundary}}
&=
\operatorname{WeightedCE}(\hat y_b,y_b)
+
0.25\,
\operatorname{SmoothL1}(\hat d_b,\tilde d_b),
\\
\mathcal L_{\mathrm{context}}
&=
0.5\,
\operatorname{MSE}
(\hat{\mathbf y}_c,\tilde{\mathbf y}_c),
\end{align}

Here distance targets are standardized after the $\log(1+d)$ transform; boundary classification uses inverse-frequency class weights. During scene-level pretraining, we optimize the six-layer scene encoder and the task-specific prediction heads. All object-encoder parameters are frozen except the final dual-stream block, which is fine-tuned at $1\times10^{-5}$. After the two-stage geometry pretraining, the complete vector geometry encoder is frozen for subsequent GeoVec-VLM training. The detailed architecture and optimization settings are reported in Table~\ref{tab:geometry-pretraining-config}.

\begin{table}[H]
\centering
\small
\caption{\textbf{Configuration of vector geometry encoder pretraining.}}
\label{tab:geometry-pretraining-config}

\setlength{\tabcolsep}{6pt}
\renewcommand{\arraystretch}{1.12}

\begin{tabularx}{\linewidth}{
@{}>{\raggedright\arraybackslash}p{0.20\linewidth}
>{\raggedright\arraybackslash}X
>{\raggedright\arraybackslash}X@{}
}
\toprule
\textbf{Setting}
& \textbf{Object level}
& \textbf{Scene level} \\
\midrule

\multicolumn{3}{@{}l}{\textit{Architecture}} \\[2pt]

Input
& 18-D element features
& 128-D object element states \\

Projection
& Linear $18\!\rightarrow\!128$
& Spatial $4\!\rightarrow\!128\!\rightarrow\!128$ \\

Encoder
& $d=128$; 4 heads;

FFN $128\!\rightarrow\!512\!\rightarrow\!128$
& $d=128$; 4 heads;

FFN $128\!\rightarrow\!512\!\rightarrow\!128$ \\

Cross-stream MLP
& \shortstack[l]{
Path: $512\!\rightarrow\!128\!\rightarrow\!128$\\
RP: $384\!\rightarrow\!128\!\rightarrow\!128$}
& -- \\

Context limit
& 256 tokens / parcel
& 384 tokens / parcel; 16 parcels; 1024 tokens / scene \\

Output
& 128-D element states; 

128-D parcel representation
& 128-D element states; 

128-D scene representation \\

\addlinespace[3pt]
\midrule
\multicolumn{3}{@{}l}{\textit{Optimization}} \\[2pt]

Optimizer
& AdamW
& AdamW \\

Learning rate
& $3\times10^{-4}$
& $3\times10^{-4}$ \\

Regularization
& weight decay 0.01; dropout 0.1; clip 1.0
& weight decay 0.01; dropout 0.1; clip 1.0 \\

Epochs
& 8
& 6 \\

Batch size / GPU
& 512
& 32 \\

\bottomrule
\end{tabularx}
\end{table}

\subsection{Spatial States, Structured Supervision, and GeoVec-VLM Training}
\label{app:GeoVec-Training}

\paragraph{Object and role states.}
For each parcel $i$, GeoVec-VLM maintains an object state $O_i$ that attends only to itself,
the geometry representations $\bar G_i$ associated with the parcel, and the visual
representations $\bar V_i$ overlapping its footprint. This restricted interaction associates vector geometry and visual content around the same spatial object while keeping the element-level geometry accessible. Given the object-selection distribution $\pi^{(r)}$ defined in
Section~\ref{sec:reasoning}, each query role $r$ reads the geometry of its selected parcel
and updates its role state. The same selection is reused across reasoning layers, while
the geometry representations continue to evolve with the backbone:
{\small
\setlength{\jot}{2pt}
\begin{equation}
\begin{aligned}
O_i &=
\operatorname{LN}\!\left(
O_i^0+
\operatorname{MHA}
\left(
O_i^0,
[O_i^0;\bar G_i;\bar V_i],
[O_i^0;\bar G_i;\bar V_i]
\right)
\right), \\
a_{rij}^{(\ell)}
&=
\operatorname{softmax}_{j}
\frac{
\left(W_{gq}\operatorname{LN}(O_i^{(\ell)})\right)^\top
W_{gk}\operatorname{LN}(g_{i,j}^{(\ell)})
}{
\sqrt{d_g}
}, \\
h_r^{(\ell)}
&=
\sum_i \pi_i^{(r)}
\sum_{j=1}^{M_i}
a_{rij}^{(\ell)}
W_{gv}\operatorname{LN}(g_{i,j}^{(\ell)}), \\
A_r^{(\ell)}
&=
\operatorname{LN}
\left(
\widetilde A_r^{(\ell)}
+
\tanh(\alpha_A)W_A h_r^{(\ell)}
\right).
\end{aligned}
\end{equation}
}%
The last equation gives the implementation of $F_A$ in Section~\ref{sec:reasoning}. It allows each role state to remain tied to the queried object while repeatedly accessing its element-level geometry.

\paragraph{Relation states.}
Relation states combine the participating role states with geometry organized according to the queried relation. For ordered relations, including direction and parcel topology, we preserve argument order through the projected role-conditioned geometry and their signed difference:
{\small
\setlength{\jot}{2pt}
\begin{equation}
\begin{aligned}
\phi_r^{(\ell)}
&=
W_P\operatorname{LN}(h_r^{(\ell)}), \\
m_e^{(\ell)}
&=
\operatorname{LN}
\left[
\phi_{r_1}^{(\ell)};
\phi_{r_2}^{(\ell)};
\phi_{r_2}^{(\ell)}-\phi_{r_1}^{(\ell)};
\left(\phi_{r_2}^{(\ell)}-\phi_{r_1}^{(\ell)}\right)
\odot
\left(\phi_{r_2}^{(\ell)}+\phi_{r_1}^{(\ell)}\right)
\right].
\end{aligned}
\end{equation}
}%
For comparative proximity, geometry is compared before object-level aggregation. For two boundary elements, let $\tilde g_a$ and $\tilde g_b$ denote their projected element-level geometry representations. Boundary-element pairs are represented using 
$\tilde g_a+\tilde g_b$,
$|\tilde g_a-\tilde g_b|$, and
$\tilde g_a\odot\tilde g_b$.
A learned scalar function scores these pairs, and a differentiable soft minimum summarizes the closest boundary interactions between the anchor and each candidate. The resulting relation representation is combined with the participating role states and written into the relation-state block $Z_e$:
{\small
\setlength{\jot}{2pt}
\begin{equation}
\begin{aligned}
B_e^{(\ell)}
&=
\left[
A_{r_1}^{(\ell)}+p_1;
\ldots;
A_{r_k}^{(\ell)}+p_k;
W_m m_e^{(\ell)}+e_{\tau_e}
\right], \\
\Delta Z_e^{(\ell)}
&=
\operatorname{MHA}
\left(
\operatorname{LN}(\widetilde Z_e^{(\ell)}),
\operatorname{LN}(B_e^{(\ell)}),
\operatorname{LN}(B_e^{(\ell)})
\right), \\
\widehat Z_e^{(\ell)}
&=
\operatorname{LN}
\left(
\widetilde Z_e^{(\ell)}
+
\tanh(\alpha_Z)\Delta Z_e^{(\ell)}
\right), \\
Z_e^{(\ell)}
&=
\operatorname{LN}
\left(
\widehat Z_e^{(\ell)}
+
\operatorname{FFN}
\left(
\operatorname{LN}(\widehat Z_e^{(\ell)})
\right)
\right).
\end{aligned}
\end{equation}
}%
This corresponds to $F_Z$ in Section~3.4. The same role- and relation-state modules are used at layers 28, 32, and 35.

\paragraph{Structured supervision.}
For relation instance $e$, let $A_e=[A_r]_{r\in\mathcal R_e}$ collect its participating role states.
Each supervised relation instance is summarized from its role and relation states by a learned relation-type query, $q_e=\operatorname{LN}\left(q_{\tau_e}+\operatorname{MHA}\left(q_{\tau_e},
[A_e;Z_e],[A_e;Z_e]\right)\right)$. For topology, the two ordered role states separately read $Z_e$, $v_j=\operatorname{LN}\left(A_{r_j}+\operatorname{MHA}(A_{r_j},Z_e,Z_e)\right)$, and form the logits $\ell_{\mathrm{top}}=\left[s_{\mathrm{in}}(1,2),s_{\mathrm{out}},s_{\mathrm{in}}(2,1)\right]$, where $s_{\mathrm{in}}$ is order-sensitive for \emph{inside}/\emph{contains}, while
$s_{\mathrm{out}}$ is symmetric for \emph{outside}. For comparative proximity, the anchor $C$ and each candidate $X\in\{A,B\}$ form a relation-specific readout $q_{CX}
=
\operatorname{LN}\left(
q_{\mathrm{dist}}
+
\operatorname{MHA}
\left(
q_{\mathrm{dist}},
[A_C;A_X;Z_{CX}],
[A_C;A_X;Z_{CX}]
\right)
\right)$, and $d_{CX}=f_{\mathrm{dist}}(q_{CX})$.
For continuous displacement, a shared two-dimensional reader takes the
relation-query state, vector-query state, and eight-way direction
probabilities. For the two object orders, its outputs are antisymmetrized as $\hat{\boldsymbol{\delta}}_{AB}
=\tfrac{1}{2}(f_{AB}-f_{BA})$. The target is the centroid displacement from $B$ to $A$, normalized by
the rendered window side length $s$,
$\boldsymbol{\delta}^{*}_{AB}=\Delta^{\rm m}_{AB}/s$.
Predictions are rescaled by $s$ for RMSE evaluation in meters. Direction, continuous displacement, parcel topology, and comparative proximity are supervised from these states:
{\small
\setlength{\jot}{2pt}
\begin{equation}
\begin{aligned}
\mathcal L_{\mathrm{direction}}
&=
\operatorname{CE}
\left(
W_{\mathrm{dir}}q_e+b_{\mathrm{dir}},
y_{\mathrm{dir}}
\right), \\
\mathcal L_{\mathrm{topology}}
&=
\operatorname{CE}
\left(
\ell_{\mathrm{top}},
y_{\mathrm{top}}
\right), \\
\mathcal L_{\mathrm{proximity}}
&=
\operatorname{CE}
\left(
[-d_{CA},-d_{CB}],
y_{\mathrm{prox}}
\right), \\
\mathcal L_{\mathrm{disp}}
&=
\operatorname{SmoothL1}
\left(
\hat{\boldsymbol{\delta}}_{AB},
\boldsymbol{\delta}^{*}_{AB}
\right).
\end{aligned}
\end{equation}
}%
Language-conditioned object selection is supervised by $\mathcal L_{\mathrm{select}}
=-\frac{1}{|\mathcal R|}\sum_{r\in\mathcal R}\log \pi_{y_r}^{(r)}$ while image--vector association follows Eq.~(4). Area selection and point-to-boundary distance have no corresponding structured decoder.

\begin{table}[H]
\centering
\small
\caption{Training configuration for GeoVec-VLM.}
\label{tab:training_details}
\begin{tabular}{ll}
\toprule
Setting & Value \\
\midrule
Stage-1 warmup
& 12,000 instances; 4$\times$Ada6000 GPUs; global batchsize 100; 240 updates \\
Stage-2 training
& 8$\times$Ada6000 GPUs; global batchsize 1,024; 680 updates \\
Vector--visual interaction layers & 8, 16 \\
Object formation layer & 24 \\
Spatial reasoning layers & 28, 32, 35 \\
Language--spatial fusion layers & 28, 32, 35 \\
LoRA rank / $\alpha$ / dropout & 16 / 32 / 0.05 \\
LoRA target & self-attention $q/k/v/o$ \\
LoRA learning rate
& stage 1: $2\times10^{-5}$; stage 2: $6\times10^{-5}$ \\
Spatial-module learning rate
& stage 1: $1\times10^{-4}$; stage 2: $4\times10^{-4}$ \\
Optimizer & AdamW \\
Weight decay & 0.01 \\
Gradient clipping & 1.0 \\
Precision & bf16 \\
Learning-rate schedule & linear warm-up followed by cosine decay \\
\bottomrule
\end{tabular}
\end{table}

\paragraph{GeoVec-VLM Training.}
GeoVec-VLM undergoes a two-stage training process. In stage 1, the structured spatial pathway is warmed up, using 12,000 QA instances, with 4,000 instances each for direction,
topology, and proximity. During this stage, vector--visual interaction operates at layers 8 and 16, object formation at layer 24, and the role and relation states at layers 28, 32, and 35. The geometry projector, spatial-state parameters, and structured prediction heads are optimized using the structured spatial objectives. In stage 2, the same spatial pathway remains active, while language--spatial fusion is enabled at layers 28, 32, and 35,
allowing language states to read the role and relation states.
The autoregressive language objective is then optimized alongside
the structured objectives over the full training split. Area and
Distance QA instances are excluded from the language-generation
objective throughout training. Language states access role and relation states via these fusion layers before proceeding through the language pathway for answer generation.

\begin{equation}
\begin{aligned}
\mathcal L_{\mathrm{S1}}
&=
\lambda_{\mathrm{assoc}}\mathcal L_{\mathrm{assoc}}
+
\lambda_{\mathrm{select}}\mathcal L_{\mathrm{select}}
+
\sum_{k\in\mathcal K}\lambda_k\mathcal L_k, \\
\mathcal L_{\mathrm{S2}}
&=
\mathcal L_{\mathrm{LM}}
+
\lambda_{\mathrm{assoc}}\mathcal L_{\mathrm{assoc}}
+
\lambda_{\mathrm{select}}\mathcal L_{\mathrm{select}}
+
\sum_{k\in\mathcal K}\lambda_k\mathcal L_k .
\end{aligned}
\end{equation}

where
$\mathcal K=\{\mathrm{direction},\mathrm{displacement},\mathrm{topology},\mathrm{proximity}\}$,
$\lambda_{\mathrm{assoc}}=\lambda_{\mathrm{select}}=0.5$,
and $\lambda_k=1$. Each structured objective is evaluated only on instances carrying its corresponding supervision, with unrelated objectives masked out.

During training, the geometry and vision encoders, along with the original Qwen3-VL backbone weights, are frozen. The backbone is adapted solely through LoRA on the $q$, $k$, $v$, and $o$ self-attention projections. The geometry projector, spatial-state modules, structured prediction heads, and language–spatial fusion modules are trained from scratch. Both stages utilize identical LoRA and optimizer configurations. The remaining optimization settings are reported in Table~\ref{tab:training_details}.

% =========================================================
\section{Dataset and Evaluation}
\label{app:experimental-details}

% 补充实验设置和 controlled comparison 的实现细节。

\subsection{Dataset Construction and Geographic Splits}
\label{app:dataset}

\begin{figure}[H]
\centering
\includegraphics[width=\linewidth]{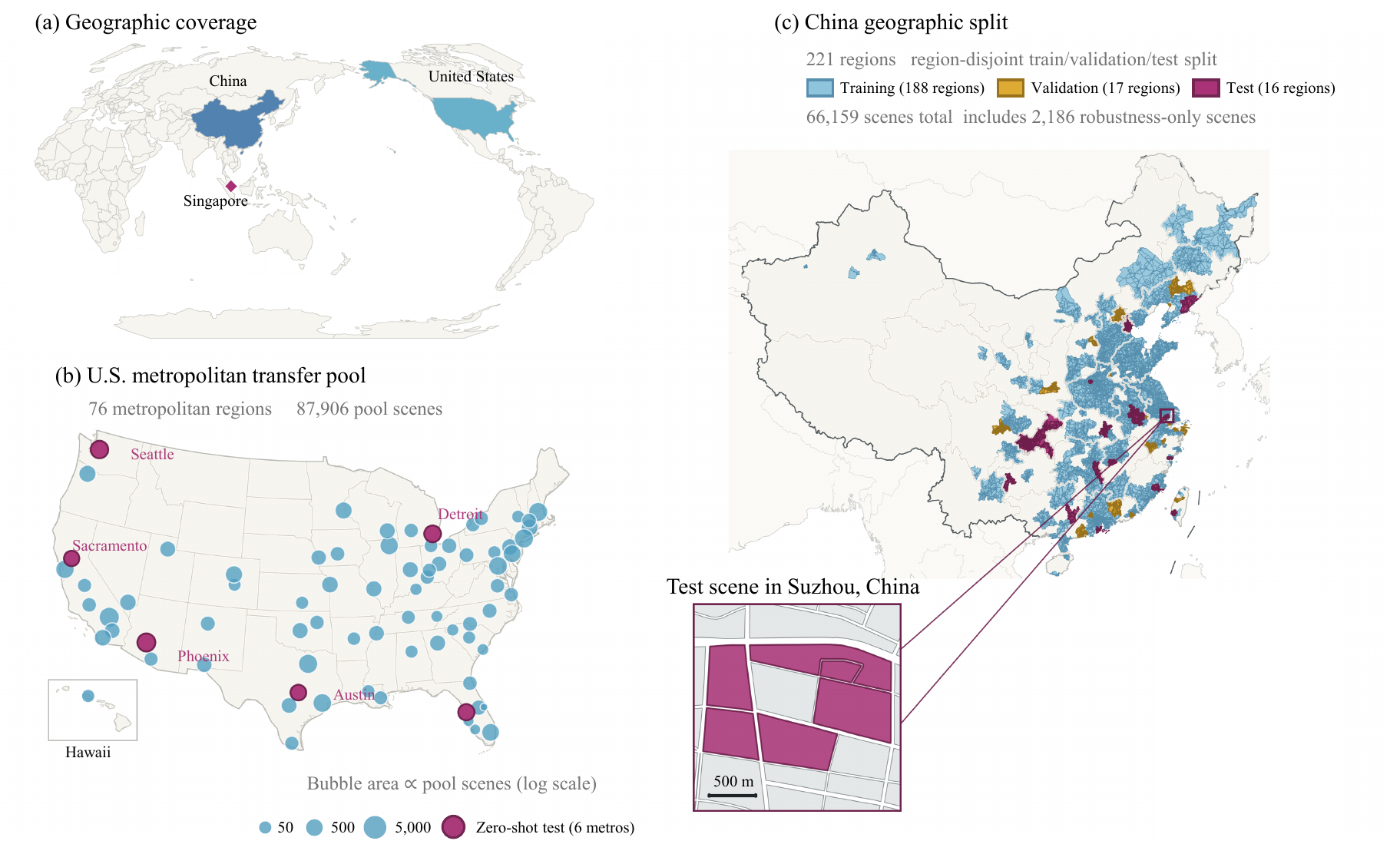}
\setlength{\abovecaptionskip}{0pt}
\caption{Geographic coverage and region-disjoint evaluation design.
(a) Geographic coverage of the China, U.S., and Singapore datasets.
(b) U.S. metropolitan regions and the six regions used for zero-shot transfer evaluation.
(c) Region-disjoint training, validation, and test split in China.}
\label{fig:data_distribution}
\end{figure}

Our data are constructed from Multi-level Street-block Cities Worldwide dataset (MSDCW) \citep{tang2025dataset}, which provides
street-block polygons at five spatial levels based on road hierarchy. We use
``parcel'' throughout the paper as a shorthand for these polygonal street-block
units. Each VLM scene pairs a \(512\times512\) image with
spatially aligned vector geometry in the same coordinate frame and can support
multiple question--answer instances.

\paragraph{Polygon preprocessing and Local context construction.}
For the China data, we use the L3 and L4 levels from 221 city-level regions.
All geometries are projected from geographic coordinates to a region-specific
metric coordinate system before metric quantities are computed, and invalid
source geometries are repaired before further processing. We then apply the
polygon preprocessing and structure-preserving boundary representation described
in Appendix~\ref{app:geometry}. To avoid retaining geometries whose shape
must be excessively simplified solely to satisfy the representation budget, we
exclude polygons for which the simplification tolerance must be increased to
meet the prescribed boundary-token cap. This filtering removes 28,715 of the
372,698 source polygons, leaving 343,983 retained polygons. 
Local spatial contexts are constructed from the retained L4 polygons separately
within each region. For each anchor parcel, neighboring parcels are selected
according to boundary proximity. We retain contexts with 3--16 neighboring
parcels, a maximum inter-parcel gap of 120\,m, and at least 0.30 facing-boundary
coverage within a 60\,m boundary buffer. This procedure produces 230,566
candidate local contexts before VLM-scene selection.

\paragraph{Image--vector scene construction.}
Each accepted context is materialized as a spatially aligned image--vector
scene. A square metric window is constructed around the retained parcels and
rendered as a deterministic north-up \(512\times512\) image, while the
corresponding vector geometries and world-to-image transform are retained.
Its resolution is fixed across scenes, whereas the geographic extent varies
with the local context; the resulting meters-per-pixel scale is therefore
recorded for each scene. Exact projected geometry is retained offline for label
construction and validation. After scene-level quality checks and materialization, the China dataset contains
66,159 scenes. Of these, 47,980, 7,998, and 7,995 scenes form the standard
training, validation, and test sets, respectively, while 2,186 additional scenes
are reserved for robustness evaluation. The downstream VLM splits are
geographically disjoint, comprising 188 training, 17 validation, and 16 test
regions. No held-out-region polygons contribute gradients during geometry-encoder
training.

% ---------------------------------------------------------
% Table 3: fixed in the text flow; not a float.
% ---------------------------------------------------------
\begin{minipage}{\linewidth}

\refstepcounter{table}
\label{tab:data-overview}

\begin{center}
\textbf{Table \thetable:} Dataset statistics for geometry pretraining, VLM training, and evaluation.
\end{center}

\vspace{-2pt}

(a) Geometry encoder data from China

\vspace{2pt}

\noindent
\begin{tabular*}{\linewidth}{
@{\extracolsep{\fill}}
lrrrr
}
\toprule
Geography
& Regions
& L3 polygons
& L4 polygons
& Context samples \\
\midrule
Training
& 188
& 65,767
& 211,139
& 185,296 \\
\midrule
Validation / Test
& 33
& 15,330
& 51,747
& 45,270 \\
\bottomrule
\end{tabular*}

\vspace{8pt}

(b) VLM scene data from China

\vspace{2pt}

\noindent
\begin{tabular*}{\linewidth}{
@{\extracolsep{\fill}}
lrrr
}
\toprule
Split & Regions & Scenes & QA instances \\
\midrule
Training   & 188 & 47,980 & 359,340 \\
Validation & 17  & 7,998  & 59,997 \\
Test       & 16  & 7,995  & 59,830 \\
\bottomrule
\end{tabular*}

\vspace{8pt}

(c) Zero-shot geographic transfer

\vspace{2pt}

\noindent
\begin{tabular*}{\linewidth}{
@{\extracolsep{\fill}}
lrrrrr
}
\toprule
Geography
& Regions
& Scenes
& Eval. regions
& Eval. scenes
& QA instances / task \\
\midrule
United States & 76 & 87,906 & 6 & 900 & 1,800 \\
Singapore     & 1  & 660    & 1 & 300 & 600 \\
\bottomrule
\end{tabular*}

\end{minipage}

\vspace{5pt}

\begin{figure}[H]
\centering
\includegraphics[width=\linewidth]{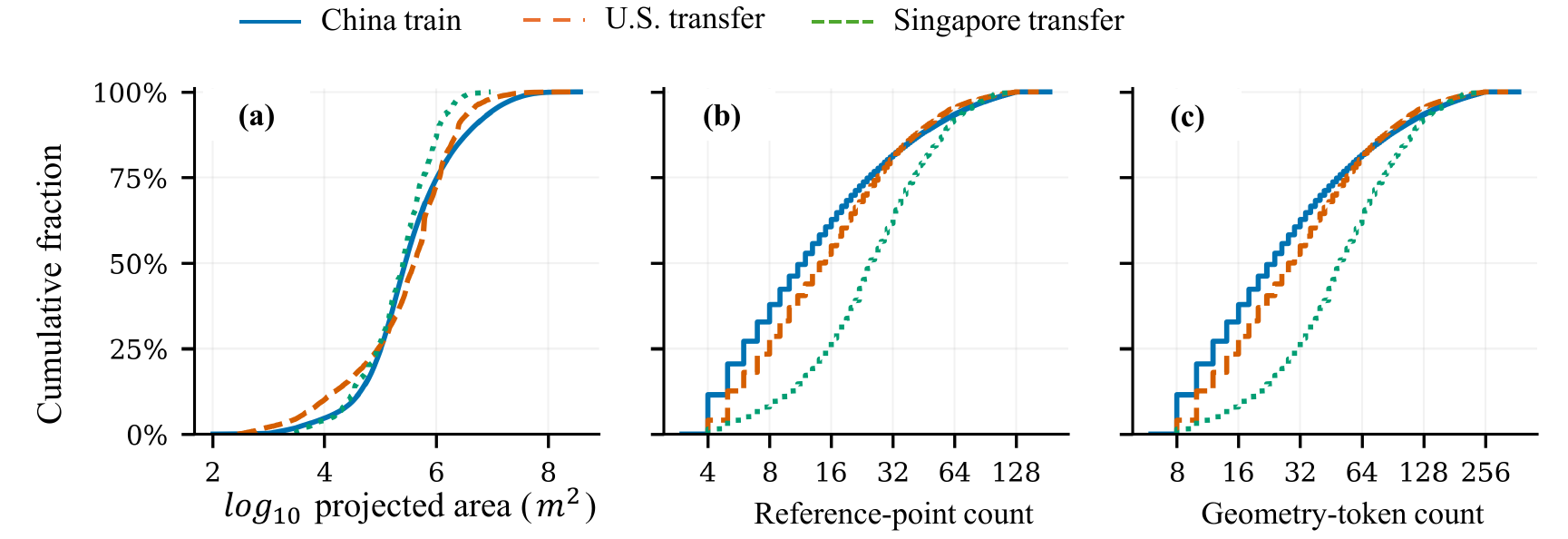}

\setlength{\abovecaptionskip}{3pt}
\caption{Geometry distribution shifts between training and transfer geographies.
Empirical cumulative distributions of (a) projected polygon area,
(b) retained reference-point count, and (c) geometry-token sequence length
for L4 polygons in the China training regions and the U.S. and Singapore
transfer geographies. Transfer distributions are computed over the polygons
in the reported evaluation regions.}
\label{fig:geometry_statistics}
\end{figure}

\paragraph{Geographic transfer.}
We further evaluate zero-shot geographic transfer to the United States and
Singapore using the same preprocessing and scene-construction pipeline. The
U.S. dataset spans 76 metropolitan regions, of which six are used for the
reported evaluation, while Singapore is evaluated as a single-region transfer
set. Neither geography contributes to model training or model selection.
Figure~\ref{fig:data_distribution} summarizes the geographic coverage and
region-disjoint evaluation design, and Table~\ref{tab:data-overview} reports
the resulting data sizes. Figure~\ref{fig:geometry_statistics} compares the scale and geometric complexity
of the L4 polygons used for China training with those from the U.S. and Singapore
transfer geographies. The U.S. polygons are moderately larger in area while
remaining similar in boundary complexity to the China training distribution.
Singapore exhibits a stronger shift in boundary complexity, with more retained
reference points and longer geometry-token sequences, despite a comparable
polygon-area range. Transfer evaluation distributions show geographic and geometric complexity variations.

\subsection{Task Construction and Metrics}
\label{app:task-construction}

\paragraph{Task selection.}
We select tasks to cover three fundamental forms of spatial reasoning
used in GIS and spatial databases: directional, topological, and metric
reasoning~\citep{guting1994introduction,carniel2024defining}.
Direction evaluates relative orientation, while parcel--parcel and
point--parcel topology test containment relations across different
spatial object types. Proximity, area, and boundary distance evaluate
metric reasoning over inter-object distance, object extent, and
point-to-boundary distance, respectively. We evaluate six spatial task families. Direction additionally uses a continuous two-dimensional displacement target on the same ordered object pairs. Table~\ref{tab:task-definitions} summarizes the evaluated relations and sample
sizes.

% ---------------------------------------------------------
% Table 4: fixed in the text flow; not a float.
% ---------------------------------------------------------
\begin{minipage}{\linewidth}
\refstepcounter{table}
\label{tab:task-definitions}

\begin{center}
\small
\textbf{Table \thetable:} Spatial tasks used in evaluation.
\end{center}

\vspace{-3pt}

\small
\noindent
\begin{tabularx}{\linewidth}{@{} l l X r @{}}
\toprule
Task & Query & Prediction & Questions \\
\midrule

Direction
& Parcel pair
& 8-way direction / continuous $(\Delta x,\Delta y)$
& 1,800 \\

Proximity
& Anchor + candidates
& Candidate with smaller boundary distance
& 1,800 \\

Parcel topology
& Parcel pair
& inside / contains / outside
& 1,800 \\

Point topology
& query point + parcel
& inside / outside
& 1,800 \\

Area
& Parcel set
& parcel with largest / smallest area
& 1,800 \\

Distance
& query point + parcel set
& parcel with nearest / farthest boundary distance to query point
& 1,800 \\

\bottomrule
\end{tabularx}

\vspace{2pt}

{\footnotesize
Question counts are reported for the China test set.
Displacement uses the same 1,800 ordered pairs as direction and therefore does
not add separate question--answer instances.
}

\end{minipage}

\vspace{6pt}

\begin{figure}[H]
\centering
\includegraphics[width=\linewidth]{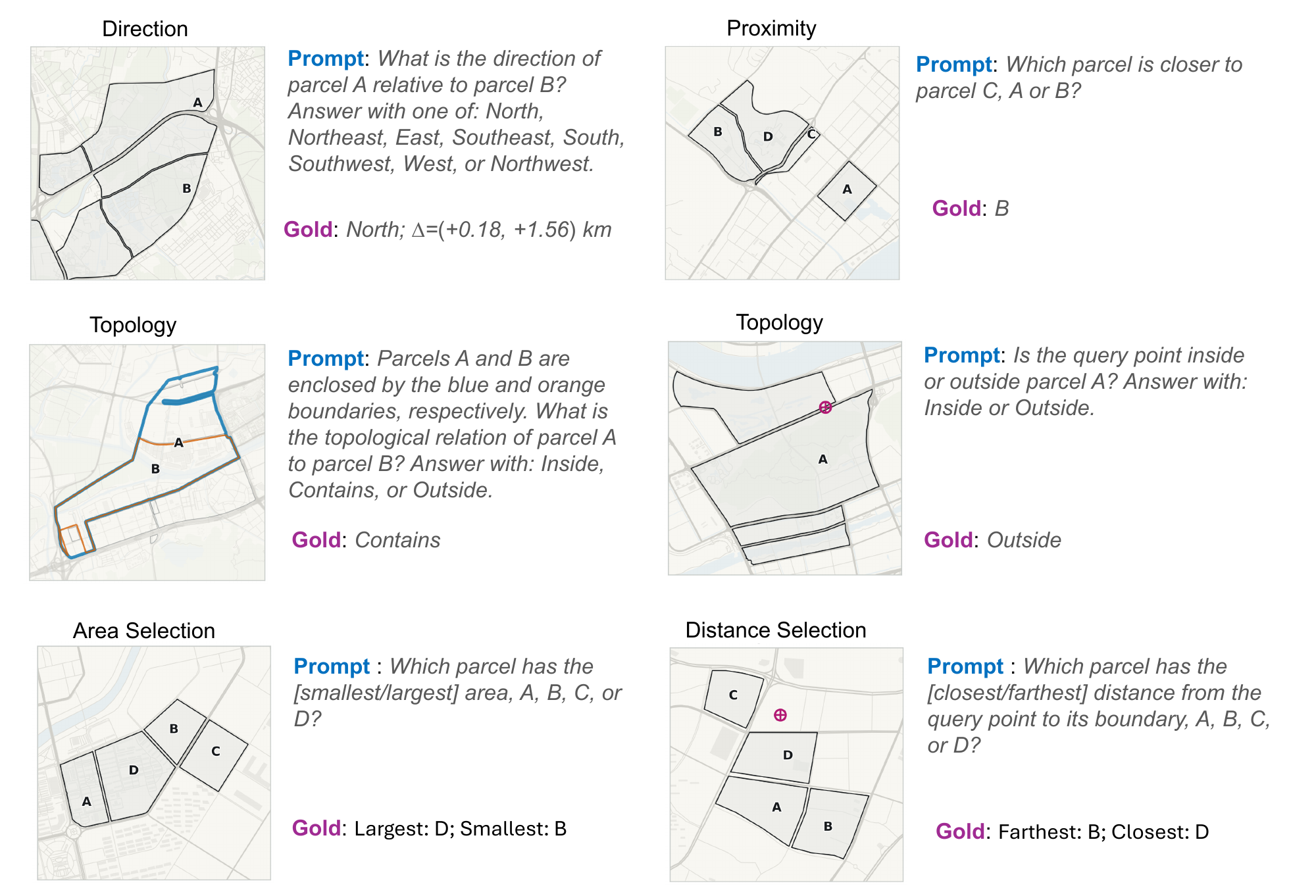}
\caption{Representative examples from the benchmark.
Each panel shows the visual input, one prompt realization, and the target answer.}
\label{fig:parcel_examples}
\end{figure}

\paragraph{Geometric target and paired query construction.}
All task labels are generated deterministically offline from the exact
projected geometries, without manual or model-generated annotation. 
Direction is defined from the ordered displacement
between parcel centroids and quantized into eight compass sectors, with the
same ordered pairs additionally assigned continuous displacement targets
$(\Delta x,\Delta y)$. Comparative proximity is determined by the minimum
polygon distance between the anchor parcel and each candidate. Parcel--parcel
and point--parcel topology are computed from exact geometric relations, while
area and boundary-distance selection use projected polygon area and
point-to-boundary distance, respectively.

To evaluate geometric reasoning beyond coarse spatial cues, we exclude
ambiguous samples and construct hard cases where commonly used geometric
proxies disagree with the exact target relation. Specifically, proximity hard cases are selected when centroid distance conflicts with minimum boundary distance; area hard cases when bounding-box area differs from polygon area; and distance hard cases when nearest-vertex distance differs from exact boundary distance. These cases reduce the possibility of solving tasks through simple geometric approximations.

Queries involving reversible or opposite relations are evaluated in matched pairs to measure geometric consistency. Direction reverses the ordered parcel pair, proximity swaps candidate objects while preserving the anchor parcel, and topology evaluates reciprocal relations such as inside/contains. Similar paired constructions are used for opposite extrema queries where applicable. These paired evaluations support the consistency analyses in Section~\ref{sec:main-results}.

Figure~\ref{fig:parcel_examples} presents representative baseline examples from the six
evaluated task families, including the visual input, one prompt realization,
and the corresponding target answer. The examples illustrate how spatial
objects, query points, and candidate sets are presented to the model. Area and
Distance include both extrema queries to evaluate metric reasoning beyond the
structured objectives used during training.

\vspace{6pt}

\begin{tcolorbox}[
    title=Common instruction for GeoVec-VLM,
    colback=white,
    colframe=blue!50,
    colbacktitle=blue!50,
    coltitle=white,
    fonttitle=\bfseries,
    boxrule=0.5pt,
    arc=1mm,
    left=5pt,
    right=5pt,
    top=5pt,
    bottom=5pt
]

\textbf{System.}
You are performing a geospatial reasoning task over a spatial scene.

\vspace{3pt}

\textbf{Object references.}
References such as \texttt{parcel\_A}, \texttt{parcel\_B}, and
\texttt{parcel\_C} denote the polygonal spatial objects assigned to the
corresponding query slots. For GeoVec-VLM, these symbols identify objects through the vector
query slots and are not rendered as text labels in the image.

\vspace{3pt}

When a query point is present, its location is defined by the center of the
crosshair marker in the image.

\vspace{3pt}

Answer using only the requested answer format.

\end{tcolorbox}

\begin{tcolorbox}[
    title=Representative task prompts,
    colback=white,
    colframe=orange!50,
    colbacktitle=orange!70,
    coltitle=white,
    fonttitle=\bfseries,
    boxrule=0.5pt,
    arc=1mm,
    left=5pt,
    right=5pt,
    top=5pt,
    bottom=5pt
]

\textbf{Direction.}

\textit{Query:}
What is the direction of \texttt{parcel\_A} relative to
\texttt{parcel\_B}?

\textit{Output:}
Answer with one of: North, Northeast, East, Southeast, South,
Southwest, West, or Northwest.

\vspace{6pt}

\textbf{Comparative proximity.}

\textit{Query:}
Which parcel is closer to \texttt{parcel\_C},
\texttt{parcel\_A} or \texttt{parcel\_B}?

\textit{Output:}
Answer with A or B.

\vspace{6pt}

\textbf{Parcel--parcel topology.}

\textit{Query:}
What is the topological relation of \texttt{parcel\_A} to
\texttt{parcel\_B}?

\textit{Output:}
Answer with one of: Inside, Contains, or Outside.

\vspace{6pt}

\textbf{Point--parcel topology.}

\textit{Query:}
Is the center of the query crosshair inside or outside
\texttt{parcel\_A}?

\textit{Output:}
Answer with Inside or Outside.

\vspace{6pt}

\textbf{Area and boundary-distance selection.}

The retained candidates are identified by letters
\texttt{A}, \texttt{B}, \texttt{C}, and \texttt{D}. The query asks for the
candidate with the largest/smallest area or nearest/farthest boundary
distance, and the output is the corresponding candidate letter.

\end{tcolorbox}

\paragraph{Prompt construction and object references.}
Each question combines a common geospatial reasoning instruction with a
task-specific query. Queried spatial objects are internally represented by
vector-object IDs and mapped to temporary symbolic slots such as
\texttt{parcel\_A}, \texttt{parcel\_B}, and \texttt{parcel\_C} during language
interaction. These slots refer to underlying vector objects and are not visual
annotations rendered in the image. For point-based tasks, the query
point corresponds to the center of the crosshair marker.

GeoVec-VLM receives images without alphabetical parcel labels, while
image-only baselines use rendered labels to identify referenced parcels.
Thus, object correspondence is provided through vector structure only for
GeoVec-VLM and through visual annotations for image-only baselines.

Training prompts are generated from compositional templates rather than a single fixed
form. Each prompt is generated by combining role descriptions, question
formulations, and output constraints. Four role descriptions, four question
formulations, and two output constraints are combined for each task family,
resulting in $4\times4\times2=32$ surface forms. Evaluation uses independently constructed prompt
forms that do not overlap with training templates.

\paragraph{Cross-modal spatial grounding metrics.}
For CSG, Object Top-1 is correct when the highest-scoring parcel
candidate (including \textit{outside}) matches the ground-truth object.
For visual support, each valid parcel slot predicts a binary mask over
visual patches using
$\hat m_{ip}=\mathbbm{1}[\sigma(s_{ip})>0.5]$.
Visual-support F1 is computed from TP, FP, and FN pooled over all valid
parcel-slot--patch pairs within a sample:
\[
F_1=\frac{2TP}{2TP+FP+FN}.
\]
Target-region patch F1 uses the same measure but restricts the comparison
to the ground-truth target parcel and its annotated visual-patch mask.
Figure~4(c) reports the per-sample target-region patch F1 distribution.

\subsection{Qualitative Spatial Reasoning Cases}
\label{app:qualitative_cases}

\begin{figure}[H]
    \centering
    \includegraphics[width=\linewidth]{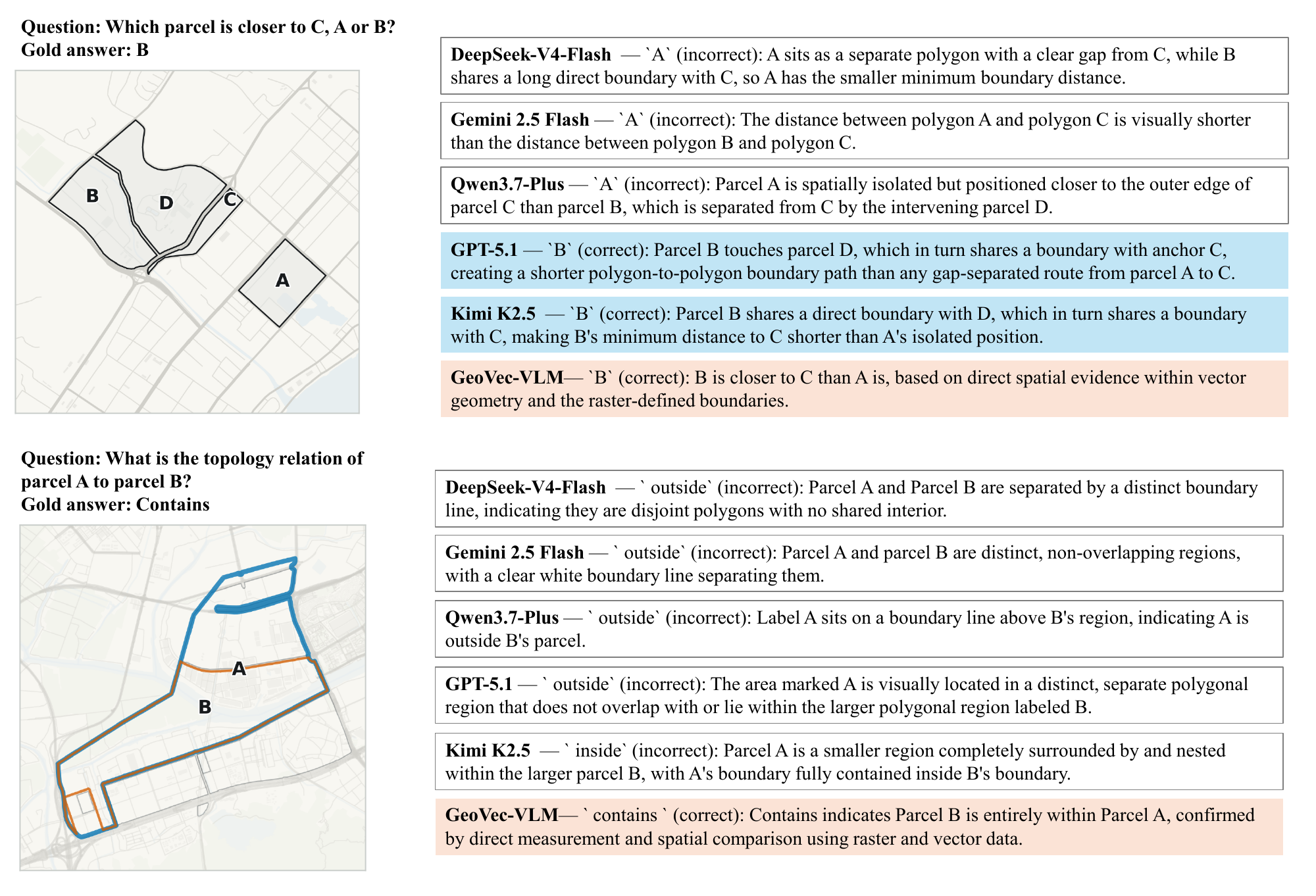}
    \caption{
    Post-hoc explanations for representative comparative proximity and parcel--parcel topology predictions. Explanations are elicited only after the constrained task answers have been collected and are used for qualitative analysis.
    }
    \label{fig:qualitative_cases}
\end{figure}

After collecting the constrained task answers used for quantitative
evaluation, we issue a separate post-hoc explanation query for
qualitative analysis. These explanations do not affect the reported
task predictions or accuracy. Figure~\ref{fig:qualitative_cases} presents two representative cases in which
fluent visual reasoning does not necessarily correspond to the underlying
geometry. In comparative proximity, several general-purpose VLMs produce plausible
visual explanations based on apparent gaps, adjacency, or intervening parcels,
yet arrive at the wrong candidate. These explanations illustrate how a model
can construct a coherent spatial narrative from the image while failing
to recover the geometric relation that determines the answer. GeoVec-VLM
selects the correct parcel and attributes the decision to spatial evidence from
the aligned vector geometry and image boundaries.
The parcel--parcel topology example exposes a related failure more directly.
Here, one parcel is contained by the other while part of their boundaries
coincide. This configuration is difficult to resolve reliably from image
appearance alone: several baselines classify the parcels as \textit{Outside},
while another reverses the ordered relation and predicts \textit{Inside}.
GeoVec-VLM instead has direct access to the underlying vector geometry in the
shared spatial frame, allowing the ordered containment relation to be resolved
from the geometric structure itself. 

\subsection{Baseline Implementation}
\label{app:geometry_baselines}

\paragraph{Spatial VLM baselines.}
SpatialRGPT and SR-3D are initialized from their released checkpoints and
adapted to our task using a balanced subset of 3,840 China-training instances
(640 per task family). We additionally tested adaptation with 12,000 training instances,
which did not improve validation performance; we therefore report
the 3,840-instance configuration. Both models receive the same image scenes and
ground-truth parcel masks through their native region-prompt interfaces.
SpatialRGPT is evaluated without depth input, while SR-3D uses its official
single-view pathway with frozen Depth Anything V2 features. Adaptation is performed with rank-8 LoRA modules on the language-model attention and MLP projections with $\alpha=16$ and zero dropout. We train for one pass over the adaptation set using answer-token cross-entropy and AdamW with a learning rate of $2\times10^{-5}$, batch size 10, 20 warm-up steps followed by cosine decay, and bf16 computation.

For API-based general-purpose VLMs, referenced parcels are marked directly in the image with alphabetical labels corresponding to the query roles. As shown in Figure~\ref{fig:parcel_examples}, these labels provide the visual reference needed to identify parcels A, B, and C; no vector geometry is supplied unless the evaluated interface explicitly includes GeoJSON.

\paragraph{Serialized GeoJSON, Geo2Vec and Poly2Vec.}
For the controlled representation comparison in Table 2, Serialized GeoJSON, Geo2Vec, and Poly2Vec use the same scene objects, query-role assignments, Qwen3-VL-8B backbone, and downstream training split.
Geo2Vec and Poly2Vec follow their respective released geometry encoding procedures and expose each object through a fixed-size embedding projected into the VLM hidden space.

Serialized GeoJSON provides vector objects as textual coordinate sequences.
Polygon and MultiPolygon geometries are transformed into a scene-local
coordinate system normalized to $[0,1]^2$ and simplified at approximately the
image resolution. The coordinate convention is explicitly specified in the
prompt, with southwest as $(0,0)$ and northeast as $(1,1)$. Each object is
represented with an opaque scene-local identifier (e.g.,
\texttt{object\_03}), while query roles are separately specified through
bindings such as \texttt{A=object\_03}. An object is serialized as
\texttt{\{"geometry":\{"coordinates":[[[0.12,0.34],...]],"type":"Polygon"\},
"properties":\{"feature\_id":"object\_00"\},"type":"Feature"\}}.

Geo2Vec and Poly2Vec replace coordinate serialization with fixed-size
object-level geometry embeddings. Geo2Vec representations are learned without
task labels using its location and shape branches, whose 128-dimensional
outputs are concatenated into a 256-dimensional embedding. Poly2Vec applies
its Fourier-based encoder to scene-normalized coordinates mapped from
$[0,1]^2$ to $[-1,1]^2$, producing a 420-dimensional embedding. These geometry
embeddings are projected into the Qwen hidden space through a two-layer MLP $h_i = W_2\,\mathrm{GELU}(W_1 f_i+b_1)+b_2$,
and the projected vectors replace the corresponding placeholder embeddings
before the VLM forward pass. The geometry embeddings remain frozen during
downstream training, while the projector and rank-16 language LoRA are
optimized under the same configuration.

% =========================================================
\section{Spatial Mechanism and Generalization}
\label{app:Spatial Mechanism and Generalization}

\subsection{Geographic Transfer Evaluation}
\label{app:Geographic Transfer}

\begin{table}[H]
\centering
\caption{
Spatial reasoning performance under geography transfer.
Baselines use image + GeoJSON; GeoVec-VLM uses native vectors.
$^\dagger$Area and Distance receive no corresponding downstream task supervision.
}
\label{tab:regional-results}

\small
\setlength{\tabcolsep}{5.2pt}
\renewcommand{\arraystretch}{1.03}

\begin{tabular}{@{}llccccc@{}}
\toprule
Region
& Model
& Direction $\uparrow$
& Proximity $\uparrow$
& Topology $\uparrow$
& Area$^\dagger$ $\uparrow$
& Distance$^\dagger$ $\uparrow$ \\
\midrule

\multirow{5}{*}{U.S.}
& DeepSeek-V4-Flash
& 0.652 & 0.591 & 0.548 & 0.758 & 0.408 \\
& Kimi K2.5
& 0.695	 & \underline{0.751} & 0.599 & \underline{0.781} & 0.590 \\
& Gemini 2.5 Flash
& 0.672 & 0.708 & 0.578 & 0.773 & 0.569 \\
& Qwen 3.7 Plus
& \underline{0.744} & 0.738 & 0.618 & 0.779 & \textbf{0.692} \\
& GPT-5.1
& 0.654 & 0.747 & \underline{0.664} & 0.738 & 0.579 \\
& \textbf{GeoVec-VLM-8B}
& \textbf{0.801} & \textbf{0.799} & \textbf{0.824}
& \textbf{0.794} & \underline{0.657} \\
\midrule

\multirow{5}{*}{Singapore}
& DeepSeek-V4-Flash
& 0.635 & 0.597 & 0.582 & 0.835 & 0.403 \\
& Kimi K2.5
& 0.667 & 0.743 & 0.634 & \underline{0.851} & \underline{0.598} \\
& Gemini 2.5 Flash
& 0.628 & 0.713 & 0.628 & \textbf{0.865} & 0.588 \\
& Qwen 3.7 Plus
& \underline{0.692} & 0.713 & 0.628 & 0.832 & \textbf{0.658} \\
& GPT-5.1
& 0.615 & \underline{0.772} & \underline{0.675} & 0.788 & 0.535 \\
& \textbf{GeoVec-VLM-8B}
& \textbf{0.775} & \textbf{0.782} & \textbf{0.848}
& 0.850 & \textbf{0.658} \\
\bottomrule
\end{tabular}
\end{table}

Table~\ref{tab:regional-results} shows that the gains in Table~\ref{tab:main-results} are not concentrated in a single test region. GeoVec-VLM ranks first on all three supervised tasks in both the U.S. and Singapore, with particularly stable gains on Topology (+16.0 and +17.3 points over the strongest image+GeoJSON baseline). Its performance on these tasks also varies by only 1.7--2.6 percentage points between the U.S. and Singapore transfer sets. Area and Distance, which receive no corresponding structured supervision, show smaller margins but remain competitive in both regions. The regional split therefore supports both the geographic robustness of GeoVec-VLM and the transferability of its learned spatial representations.

\subsection{Targeted Geometry-Token Interventions}
\label{app:geometry_intervention}

We further examine which components of the structured geometry representation
support spatial reasoning. Starting from the trained GeoVec-VLM, we apply three
targeted interventions at inference time while keeping the model parameters,
visual and language inputs, and non-targeted geometry cues unchanged. 
As shown in Table~\ref{tab:geometry_intervention_design}, the three
interventions remove the shared scene reference (\textsc{-SceneRef}), suppress
reference-point geometry (\textsc{-RPGeom}), or remove explicit hierarchical
identifiers (\textsc{-HierID}). Table~\ref{tab:geometry_intervention} reports
the resulting change in final-answer accuracy relative to the intact model.

\begin{table}[H]
\centering
\caption{
Definition of the targeted geometry interventions.
Each intervention disrupts one source of geometric information while
preserving the remaining representation.
}
\label{tab:geometry_intervention_design}

\small
\setlength{\tabcolsep}{5pt}
\renewcommand{\arraystretch}{1.08}

\begin{tabularx}{\linewidth}{
    @{}
    l
    >{\RaggedRight\arraybackslash}X
    >{\RaggedRight\arraybackslash}X
    @{}
}
\toprule
Intervention & Disrupted information & Preserved information \\
\midrule

\textsc{-SceneRef}
&
Object-level H3/centroid placement and RP tile/context coordinates
in the shared scene frame.
&
Within-object relative contour, path attributes, hierarchical
identifiers, and object identity.
\\[2pt]

\textsc{-RPGeom}
&
Object-local RP coordinates, bearing/turn features, and pointwise
tile/local/context coordinates, which are collapsed to the object center.
&
Object centers and cross-object placement, arc phase, path attributes,
and hierarchical identifiers.
\\[2pt]

\textsc{-HierID}
&
Explicit part, ring, ring-role, local-structure, and phase identifiers.
&
Token type, sequence order, RP/path alternation, attention masks,
and boundary-neighbor connectivity.
\\

\bottomrule
\end{tabularx}
\end{table}

\begin{table}[H]
\centering
\caption{\textbf{Effects of targeted geometry-token interventions across evaluation regions.}
Values are percentage-point changes in final-answer accuracy relative to the intact GeoVec-VLM. Larger negative changes indicate greater reliance on the disrupted spatial information}
\label{tab:geometry_intervention}

\small
\renewcommand{\arraystretch}{1.05}

\begin{tabular*}{\textwidth}{
@{\extracolsep{\fill}}
llrrrrrr
@{}
}
\toprule
Geography & Intervention
& Dir. $\Delta$
& Prox. $\Delta$
& Parcel Topo. $\Delta$
& Point Topo. $\Delta$
& Area $\Delta$
& Dist. $\Delta$ \\
\midrule
\addlinespace[2pt]

China
& $-$SceneRef
& -63.94 & -30.06 & -20.06 & -0.67 & -0.61 & -23.61 \\
& $-$RPGeom
& +1.39 & -4.61 & -17.50 & -0.56 & -4.28 & +0.06 \\
& $-$HierID
& -13.06 & -8.72 & -8.72 & -7.17 & -3.83 & -1.94 \\

\midrule
\addlinespace[2pt]

U.S.
& $-$SceneRef
& -61.39 & -28.11 & -33.72 & -1.11 & -4.00 & -26.72 \\
& $-$RPGeom
& -2.72 & -4.11 & -17.17 & -0.33 & -7.17 & +0.39 \\
& $-$HierID
& -10.78 & -22.59 & -10.78 & -12.94 & -8.39 & -6.17 \\

\midrule
\addlinespace[2pt]

Singapore
& $-$SceneRef
& -60.83 & -26.50 & -37.83 & -2.50 & -4.50 & -28.00 \\
& $-$RPGeom
& -0.17 & -4.83 & -9.67 & -2.67 & -7.50 & -3.17 \\
& $-$HierID
& -27.17 & -14.33 & -18.00 & -6.76 & -11.50 & -12.00 \\

\bottomrule
\end{tabular*}

\end{table}

The three interventions reveal distinct roles of the geometric information
preserved by GeoVec. Removing global placement produces the largest degradation
on Direction and Proximity, and also substantially affects Parcel Topology and
Distance. These tasks depend on relative object positions or cross-object metric
relations in a shared coordinate frame. In contrast, removing local shape has
little effect on Direction but consistently degrades Parcel Topology and Area,
showing that within-object contour information contributes more directly to
shape- and boundary-sensitive reasoning. Removing explicit structure IDs causes
broader but generally smaller degradation, indicating that hierarchical
identifiers help organize element-level geometry beyond the information already
available from token order, RP/path alternation, and boundary connectivity.

These intervention effects vary across geographies.
For example, the Parcel Topology drop caused by removing global placement
increases from 20.1 points in China to 33.7 in the U.S. and 37.8 in Singapore,
while removing structure IDs has larger effects on several tasks in the transfer sets.
These changes are consistent with the geometry-distribution differences
reported in Fig.~\ref{fig:geometry_statistics}, including shifts in polygon scale and boundary
complexity. The intervention results therefore suggest that the same structured
representation can support a task through different mixtures of geometric cues
as the underlying parcel distribution changes.

\begin{figure}[H]
\centering
\includegraphics[width=\linewidth]{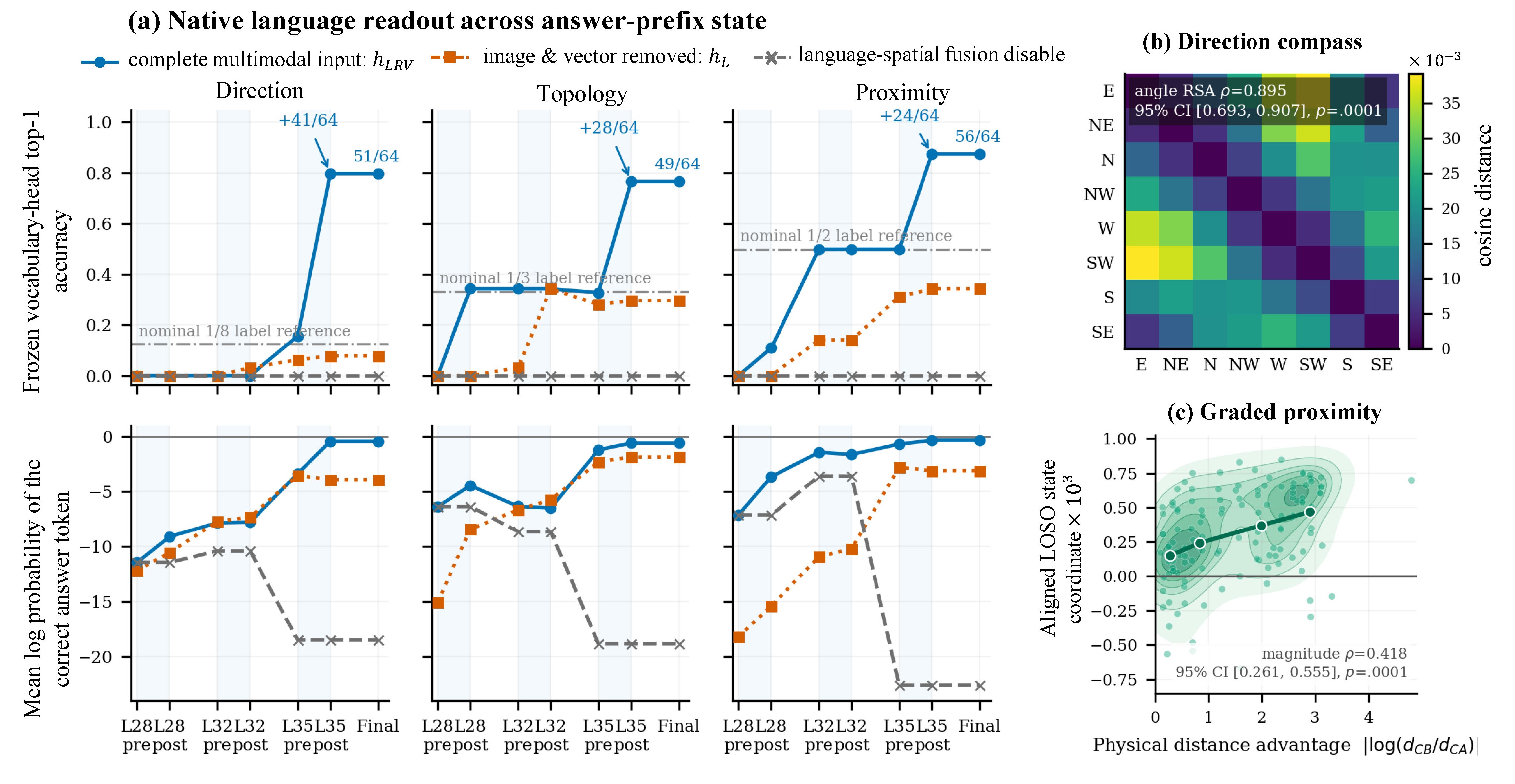}
\caption{\textbf{Spatial information enters and structures the language path.}
(a) Vocabulary-head readout of the answer-prefix state around language--spatial fusion. 
With multimodal inputs, top-1 accuracy rises sharply after the language states read the spatial memory; removing this fusion suppresses the transition.
(b) Pairwise distances between direction-conditioned language states increase with physical angular separation.
(c) Proximity-state variation tracks the physical distance advantage between candidates.}
\label{fig:answer-states}
\end{figure}

\subsection{How Vector Geometry Influences Answer Generation}
\label{app:language_spatial_fusion}

Figure~\ref{fig:answer-states}(a) examines how instance-specific spatial
information enters the language path. Without image and vector inputs,
correct-answer probability may increase across language layers without
improving top-1 accuracy. With complete inputs, accuracy rises sharply after
language--spatial fusion; removing the fusion eliminates this transition.
This identifies the spatial states as the pathway through which
instance-specific spatial information reaches answer generation.
The language states after fusion also preserve continuous geometric structure shown in Fig.~\ref{fig:answer-states}(b--c): representational distance tracks angular separation for direction ($\rho=0.895$) and the physical distance advantage for proximity ($\rho=0.418$). The fused language representation therefore retains geometric structure beyond the discrete task labels.

% =========================================================
\section{Efficiency and Capability Preservation}
\label{app:Efficiency and Capability Preservation}

\subsection{Input token consumption.}
\label{app:input token consumption}

\begin{figure}[H]
\centering
\includegraphics[width=\linewidth]{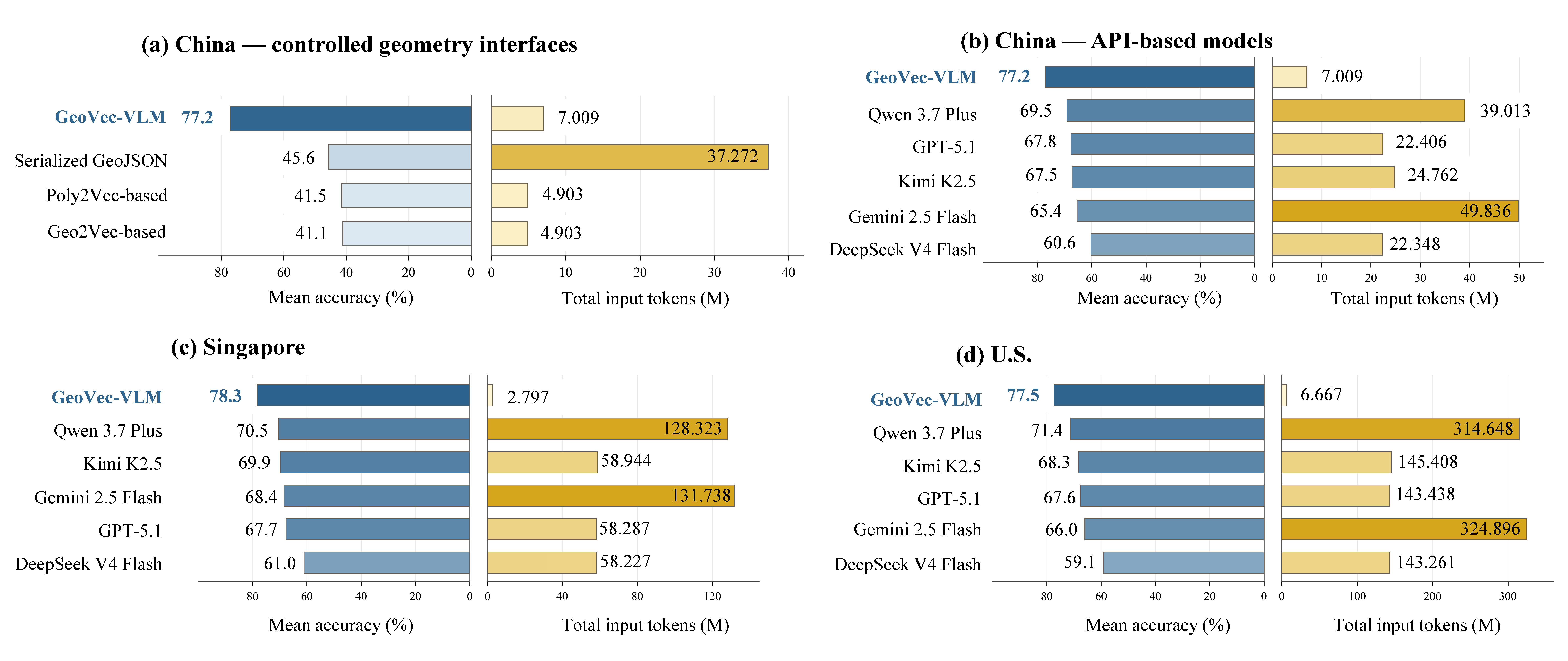}
\setlength{\abovecaptionskip}{3pt}
\caption{Total input-token consumption and spatial reasoning accuracy across evaluation sets. Token counts include the complete model input, including visual inputs, language instructions, and geometry when present. For API models, counts are those reported by the corresponding model interface. China and U.S. each contain 10,800 evaluated questions, while Singapore contains 3,600. Geo2Vec and Poly2Vec are evaluated only in the controlled China setting, consistent with Table ~\ref{tab: geometry inferace}.}
\label{fig:total input tokens}
\end{figure}

Fig.~\ref{fig:total input tokens} compares the total input consumed over each evaluation set with the corresponding mean task accuracy. On the controlled China benchmark, Geo2Vec and Poly2Vec use 4.90M input tokens but expose each vector object through fixed-size embeddings; increasing the input to 7.01M with element-level GeoVec tokens raises mean accuracy from 41.1–41.5\% to 77.2\%. Serialized GeoJSON exhibits the opposite regime, consuming 37.27M tokens while reaching only 45.6\% mean accuracy. The same pattern becomes more pronounced for API models on the transfer sets, where GeoJSON-based inputs consume 143–325M tokens in the U.S. and 58–132M in Singapore, substantially exceeding GeoVec while remaining less accurate. These results suggest that the useful bandwidth of a geometry interface depends on both how much geometric information remains accessible and how that information is structurally organized for multimodal computation.

\subsection{General Capability Preservation}
\label{app:General Capability Preservation}

\begin{table}[H]
\centering
\caption{\textbf{General capability after GeoVec adaptation.}
GeoVec retains the trained language parameters while disabling the
GeoVec-specific spatial pathway.}
\label{tab:general_capability}

\small
\renewcommand{\arraystretch}{1.05}

\begin{tabular*}{\textwidth}{
@{\extracolsep{\fill}}
llrrrr
@{}
}
\toprule
Benchmark & Scope & $N$ & Qwen & GeoVec & $\Delta$ (pp) \\
\midrule
\addlinespace[2pt]

IFEval
& Overall
& 541
& 84.47\%
& 85.03\%
& +0.55 \\

\midrule
\addlinespace[2pt]

MMLU
& Overall
& 1140
& 77.54\%
& 77.28\%
& -0.26 \\

& STEM
& 360
& 73.06\%
& 72.78\%
& -0.28 \\

& Humanities
& 260
& 80.77\%
& 80.38\%
& -0.38 \\

& Other
& 280
& 76.79\%
& 76.79\%
& +0.00 \\

& Social sciences
& 240
& 81.67\%
& 81.25\%
& -0.42 \\

\bottomrule
\end{tabular*}
\end{table}

GeoVec-VLM adapts the frozen Qwen3-VL-8B-Instruct backbone with LoRA during
spatial training. We examine whether this adaptation alters capabilities of the
original backbone that are unrelated to the spatial tasks used for training.
We retain the trained language parameters while disabling the GeoVec-specific
spatial modules and evaluate instruction following on IFEval
\citep{zhou2023instruction} and multidisciplinary knowledge on MMLU
\citep{hendrycks2021measuring}.

Table~\ref{tab:general_capability} shows little change relative to the original
Qwen3-VL-8B-Instruct. Strict prompt accuracy on IFEval changes from 84.47\% to
85.03\%, while MMLU accuracy changes from 77.54\% to 77.28\%. The MMLU
differences remain small across STEM, humanities, social sciences, and other
subjects. The paired 95\% confidence intervals for IFEval
($[-1.29,+2.40]$ percentage points) and MMLU
($[-0.61,+0.09]$) both include zero. These diagnostics provide no evidence of
a measurable loss in instruction following or multidisciplinary knowledge
after GeoVec training.

\end{document}